\documentclass[acmsmall]{acmart}
\usepackage{makecell}
\usepackage{listings}
\usepackage{subcaption}
\usepackage{xcolor}
\usepackage{multirow}
\usepackage{verbatim}
\usepackage{framed}

\definecolor{codegreen}{rgb}{0,0.6,0}
\definecolor{codegray}{rgb}{0.5,0.5,0.5}
\definecolor{codepurple}{rgb}{0.58,0,0.82}
\definecolor{backcolour}{rgb}{0.95,0.95,0.92}

\lstdefinestyle{mystyle}{
    backgroundcolor=\color{backcolour},   
    commentstyle=\color{codegreen},
    keywordstyle=\color{magenta},
    numberstyle=\tiny\color{codegray},
    stringstyle=\color{codepurple},
    basicstyle=\ttfamily\footnotesize,
    breakatwhitespace=false,         
    breaklines=true,                 
    captionpos=b,                    
    keepspaces=true,                 
    numbers=left,                    
    numbersep=5pt,                  
    showspaces=false,                
    showstringspaces=false,
    showtabs=false,                  
    tabsize=2
}

\AtBeginDocument{%
  }

\setcopyright{acmlicensed}
\acmJournal{CSUR}
\acmYear{2025} \acmVolume{1} \acmNumber{1} \acmArticle{1} \acmMonth{1}\acmDOI{10.1145/3737873}
\begin{document}

\title{A Survey on Employing Large Language Models for Text-to-SQL Tasks}

\author{Liang Shi}
\authornote{Both authors contributed equally to this research.}
\email{liang.shi@stu.pku.edu.cn}
\orcid{0009-0009-3726-0299}
\author{Zhengju Tang}
\authornotemark[1]
\email{zhengju.tang@stu.pku.edu.cn}
\orcid{0009-0001-9299-2662}
\affiliation{%
  \institution{School of Computer Science, Peking University}
  \streetaddress{5 Yiheyuan Road}
  \city{Beijing}
  \postcode{100871}
  \country{China}
}

\author{Nan Zhang}
\affiliation{%
  \institution{SINGDATA CLOUD PTE. LTD}
  \city{Bellevue}
  \country{USA}}
\email{eric_zhn@hotmail.com}
\orcid{0009-0006-3678-5059}

\author{Xiaotong Zhang}
\affiliation{%
  \institution{SINGDATA CLOUD PTE. LTD}
  \city{Beijing}
  \country{China}}
\email{meeme9@163.com}
\orcid{0009-0000-2725-2927}

\author{Zhi Yang}
\authornote{Zhi Yang is the corresponding author.}
\affiliation{%
  \institution{School of Computer Science, Peking University}
  \city{Beijing}
  \country{China}}
\email{yangzhi@pku.edu.cn}
\orcid{0000-0002-8219-4499}

\renewcommand{\shortauthors}{Shi and Tang, et al.}

\begin{abstract}
  With the development of the Large Language Models (LLMs), a large range of LLM-based Text-to-SQL(Text2SQL) methods have emerged. This survey provides a comprehensive review of LLM-based Text2SQL studies. We first enumerate classic benchmarks and evaluation metrics. For the two mainstream methods, prompt engineering and finetuning, we introduce a comprehensive taxonomy and offer practical insights into each subcategory. We present an overall analysis of the above methods and various models evaluated on well-known datasets and extract some characteristics. Finally, we discuss the challenges and future directions in this field.
\end{abstract}

\ccsdesc[500]{Computing methodologies~Natural language processing}
\ccsdesc[500]{Computing methodologies~Machine translation}
\ccsdesc[300]{Computing methodologies~Knowledge representation and reasoning}
\ccsdesc[300]{Computing methodologies~Information extraction}

\keywords{Large Language Models, Text-to-SQL, Prompt Engineering, Fine-tuning}

\received{12 November 2024}
\received[revised]{5 May 2025}
\received[accepted]{18 May 2025}

\maketitle

\section{Introduction}

In the era of big data, a significant portion of data is stored in relational databases, which serve as the backbone for various organizational data management systems. As the volume of data continues to increase, the capability to efficiently query and leverage this data has emerged as a pivotal factor in enhancing competitiveness across numerous sectors in this era. Relational databases require the use of SQL for querying. However, writing SQL necessitates specialized knowledge, which creates barriers for unprofessional users to query and access databases.


Text-to-SQL parsing is a well-established task in the field of natural language processing (NLP). Its purpose is to convert natural language queries into SQL queries, bridging the gap between non-expert users and database access. To illustrate, imagine a table named cities with three columns: city\_name (type: string), population (type: integer), and country (type: string). If we are given the natural language query "Find all the cities with a population greater than 1 million in the United States," the Text-to-SQL parsing technique should automatically generate the correct SQL query: "SELECT city\_name FROM cities WHERE population > 1000000 AND country = 'United States'." Researchers have made substantial progress in this area. Initially, template-based and rule-based methods\cite{zelle1996learning, li2014constructing} were employed. These approaches involved creating SQL templates for various scenarios. While template-based methods showed promise, they required significant manual effort. With the rapid advancement of deep learning, Seq2Seq\cite{sutskever2014sequence} methods have emerged as the mainstream approach. Seq2Seq\cite{sutskever2014sequence} models provide an end-to-end solution, directly mapping natural language input to SQL output, which eliminates the need for intermediate steps like semantic parsing or rule-based systems. Among Seq2Seq\cite{sutskever2014sequence} methods, pre-trained language models(PLMs), which serve as predecessors to large language models (LLMs), show promise in text-to-SQL tasks. Benefiting from the rich linguistic knowledge in large-scale corpora, PLMs become the state-of-the-art (SOTA) solution during that period\cite{qin2022survey}.


As model sizes and training data continue to grow, pre-trained language models (PLMs) naturally evolve into large language models (LLMs), exhibiting even greater power. Due to the scaling law\cite{kaplan2020scaling} and their emergent capabilities\cite{wei2022emergent}, LLMs have made substantial contributions across diverse domains, including chatbots\cite{dam2024complete}, software engineering\cite{hou2023large}, and agents\cite{wang2024survey}, etc. The remarkable capabilities of LLMs have prompted research into their application for text-to-SQL tasks. Current literature on LLM-based text-to-SQL mainly focuses on two main approaches of LLMs: prompt engineering and fine-tuning. Prompt engineering methods take advantage of the instruction-following capability of LLMs to implement well-designed workflows. Moreover, prompt engineering methods frequently utilize retrieval-augmented generation(RAG) and few-shot learning to obtain helpful knowledge and demonstrations, and employ reasoning techniques such as Chain-of-Thought(CoT)\cite{wei2022chain} to further enhance performance. Fine-tuning methods follow the "pre-training and fine-tuning" learning paradigm of PLMs\cite{zhao2023survey} and involve training an pretrained LLM on text-to-SQL datasets. There exists a trade-off between prompt engineering methods and fine-tuning methods. Usually, prompt engineering demands less data but may lead to suboptimal results, while fine-tuning can enhance performance but necessitates a larger training dataset.

This paper aims to provide a comprehensive survey of employing LLMs for text-to-SQL tasks and will introduce LLM-based text-to-SQL in the following aspects:
\begin{itemize}
\item \textbf{Overview}: We provide a brief introduction to LLMs and LLM-based text-to-SQL methods. We also summarize the differences between traditional and LLM-based Text-to-SQL approaches as well as the advantages of LLM-based Text-to-SQL.

\item \textbf{Benchmark and evaluation metrics}: For benchmarks, we classify them into two categories: benchmarks prior to the rise of LLM and benchmarks in the era of LLMs. We summarize the statistics and characteristics of each benchmark. Additionally, we introduce in detail five benchmarks in the era of LLMs. These datasets were created after the widespread use of LLM and present new challenges such as domain knowledge, large-scale table schemas, diverse perturbations, and real-world enterprise-level scenarios. For evaluation metrics, we introduce metrics frequently used in text-to-SQL tasks. In addition, we introduce some benchmaking studies.

\item \textbf{Prompt engineering}: We categorize the prompt engineering methods for text-to-SQL into three stages, namely pre-processing, inference, and post-processing. Pre-processing will handle the format and layout of questions and table schemas. We also emphatically introduce schema linking techniques in the pre-processing stage. In the inference stage, we explain how LLM-based text-to-SQL methods generate the corresponding SQL query when provided with the user's question and corresponding database schemas. In the final post-processing stage, we introduce how to enhance the performance and stability of LLM-based Text-to-SQL methods after generating SQL queries.

\item \textbf{Finetuning}: We organize this section into three main components, namely finetuning objectives, training methods, training data and model evaluation. Firstly, we will summarize finetuning objectives in recent LLM-based text-to-SQL papers, which include the SQL generation objective and other objectives aimed at enhancing the performance of different steps. Then frequently used training methods will be introduced. Next, there will be an introduction on how to obtain finetuning data. Finally, the model evaluation methods will be discussed.

\item \textbf{Model}: We classify LLMs according to whether they are open-source and present the applications of these models in the text-to-SQL domain. Additionally, we analyze the development trend of these LLMs, which reveals the usage frequency of the base models in the LLM-based text-to-SQL methods over time.

\item \textbf{Analysis}: We conduct a comparative analysis of experimental results from existing studies to further investigate the practical value and applicability of methods and models.

\item \textbf{Future directions}: We conduct a review of the error analysis of research approaches. Subsequently, we introduce the challenges faced by existing solutions in practical settings and provide some promising future development direction of LLM-based text-to-SQL tasks.
\end{itemize}

As one of the earliest surveys\cite{zhang2024natural, hong2024next, liu2024survey, zhu2024large} on LLM-based Text-to-SQL, our work distinguishes itself by introducing a systematic taxonomy of existing methods, providing a comprehensive analysis of the entire LLM-powered Text-to-SQL pipeline, and reinforcing practical utility through concise "Key Takeaways" at the end of each section. We intend this survey to serve as a valuable resource for newcomers to this field and offer valuable insights for researchers.


\begin{figure}
    \centering
    \includegraphics[width=0.95\textwidth]{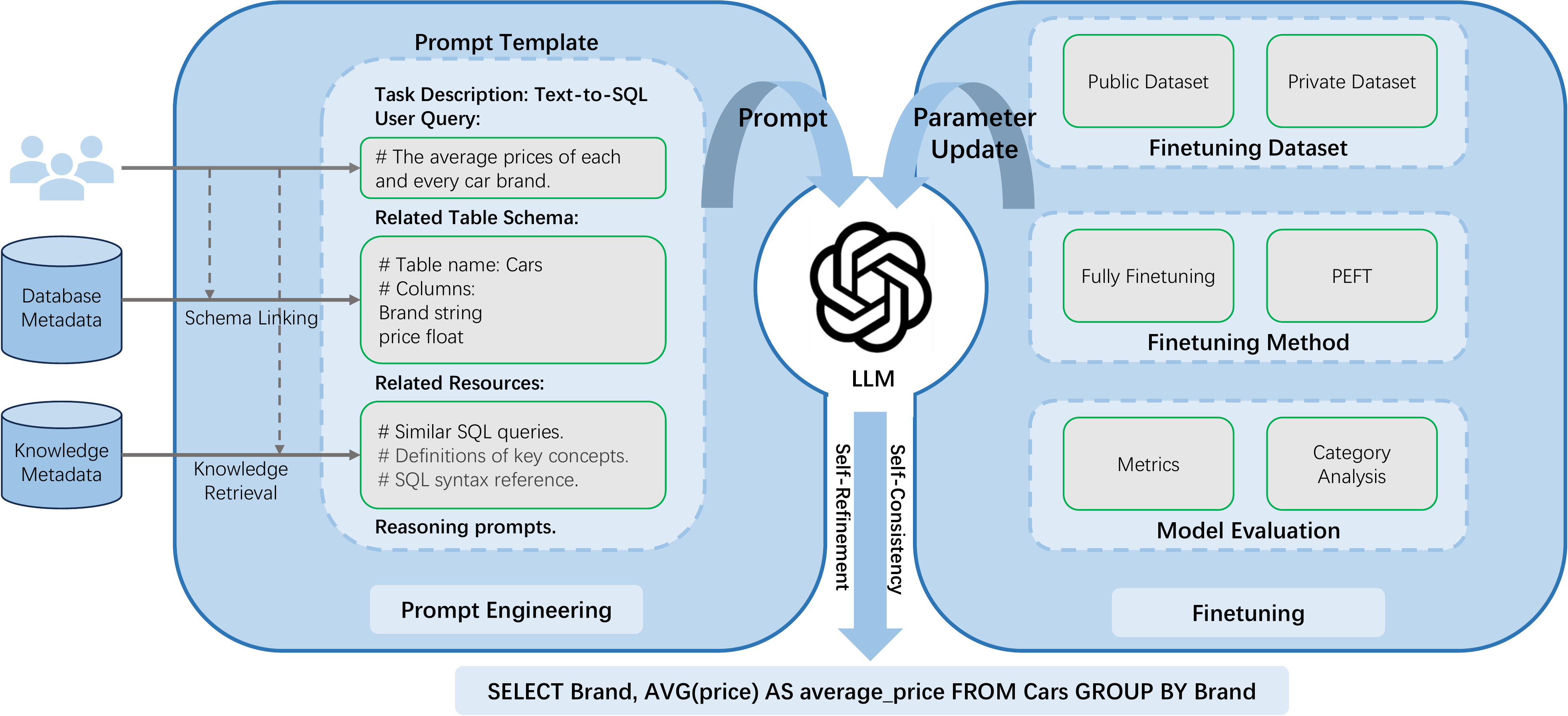}
    \caption{Framework of employing LLMs in Text-to-SQL}
    \label{fig:framework_all}
\end{figure}
\section{Overview}

\subsection{LLMs and LLM-based Text-to-SQL}

Large language models(LLMs) have emerged as a milestone for natural language processing and machine learning. The concept of LLMs comes from the practice of continually scaling up the parameter size of pretrained language models (PLMs) and the volume of training data\cite{zhao2023survey}, which results in surprising abilities, known as emergent abilities\cite{wei2022emergent}, that are not found in smaller PLMs. One example of emergent capability is few-shot learning\cite{brown2020language}, meaning that LLMs can complete the downstream tasks with several proper task demonstrations in the prompt without further training. Another example is the instruction following ability\cite{wei2021finetuned}, with which LLMs have been shown to respond appropriately to instructions describing an unseen task.

Due to the emergent abilities\cite{wei2022emergent} of LLMs and the basic operating principle of LLMs which gradually produce the next word that has the highest probability based on the input prompt\cite{zhao2023survey}, prompt engineering becomes one of the two primary streams to apply LLMs to downstream tasks. The representative approaches of prompt engineering are Retrieval Augmented Generation(RAG)\cite{gao2023retrieval}, few-shot learning\cite{brown2020language}, and reasoning\cite{wei2022chain, zhou2022least, yao2024tree}. The other stream of LLMs' application to downstream tasks is called fine-tuning, which follows the "pre-training and fine-tuning" learning paradigm of PLMs\cite{zhao2023survey} and aims to boost performance in specific domains and address privacy concerns. The general fine-tuning process mainly involves data preparation, pre-training model selection, model fine-tuning, and model evaluation.

Text-to-SQL, which is a challenging task in both natural language processing(NLP) and database communities and involves mapping natural language questions on the given relational database into SQL queries, has also been revolutionized by the emergence of LLMs. We summarize the general framework of LLM-based Text-to-SQL systems in fig.\ref{fig:framework_all}. Based on the two main streams of LLM applications, we categorize the methods used in LLM-based Text-to-SQL into two categories: prompt engineering and fine-tuning. As for prompt engineering methods, we typically design a well-structured prompt encompassing various components, such as task descriptions, table schemas, questions, and additional knowledge, and utilize in-context learning and reasoning methods at the same time. As for fine-tuning methods, we typically generate or collect text-to-SQL datasets, select appropriate pretrained LLMs and finetuning approaches such as LORA\cite{hu2021lora}, and compare the test results before and after to understand the changes in the model's performance.

\subsection{Difference between traditional and LLM-based Text-to-SQL Approaches}

Prior to the widespread of LLMs, there were two primary streams for text-to-SQL methods. One stream involves utilizing the sequence-to-sequence (Seq2Seq) model\cite{sutskever2014sequence}, in which an encoder is designed to capture the semantics of the natural language(NL) question and the corresponding table schema, and a decoder is employed to generate the SQL query token by token based on the encoded question representation. Some notable methods in this approach include IRNet\cite{jha2019irnet}, SQLNet\cite{xu2017sqlnet}, Seq2SQL\cite{WikiSQL}, HydraNet\cite{lyu2020hybrid}, Ryansql\cite{choi2021ryansql}, Resdsql\cite{li2023resdsql} and ISESL-SQL\cite{10.1145/3534678.3539294}. The other stream involves finetuning PLMs such as BERT\cite{devlin2018bert}, which leverage the extensive knowledge present in large-scale text collections and have proved to be effective in enhancing the performance of downstream text-to-SQL parsing tasks.

We argue that there are two distinct aspects that differentiate traditional text-to-SQL approaches from LLM-based ones:
\begin{itemize}
\item \textbf{Novel Paradigm}: Traditional text-to-SQL approaches necessitate training, whereas LLMs can often circumvent this requirement. Leveraging the instruction following ability\cite{wei2021finetuned} of LLMs, LLMs can accomplish the text-to-SQL tasks with the aid of appropriate instructions and information.

\item \textbf{Uniform Architecture}: According to a previous survey\cite{qin2022survey}, the encoder and decoder in traditional approaches can be designed using diverse architectures such as LSTM\cite{hochreiter1997long}, Transformer\cite{vaswani2017attention}, and even GNNs\cite{bogin2019representing}. In contrast, LLMs adhere to a uniform transformer-based architecture, which not only enables easier scaling up but also facilitates a more streamlined implementation.
\end{itemize}

\subsection{Why LLM-based Text-to-SQL}

\begin{figure}
    \centering
    \includegraphics[width=0.75\linewidth]{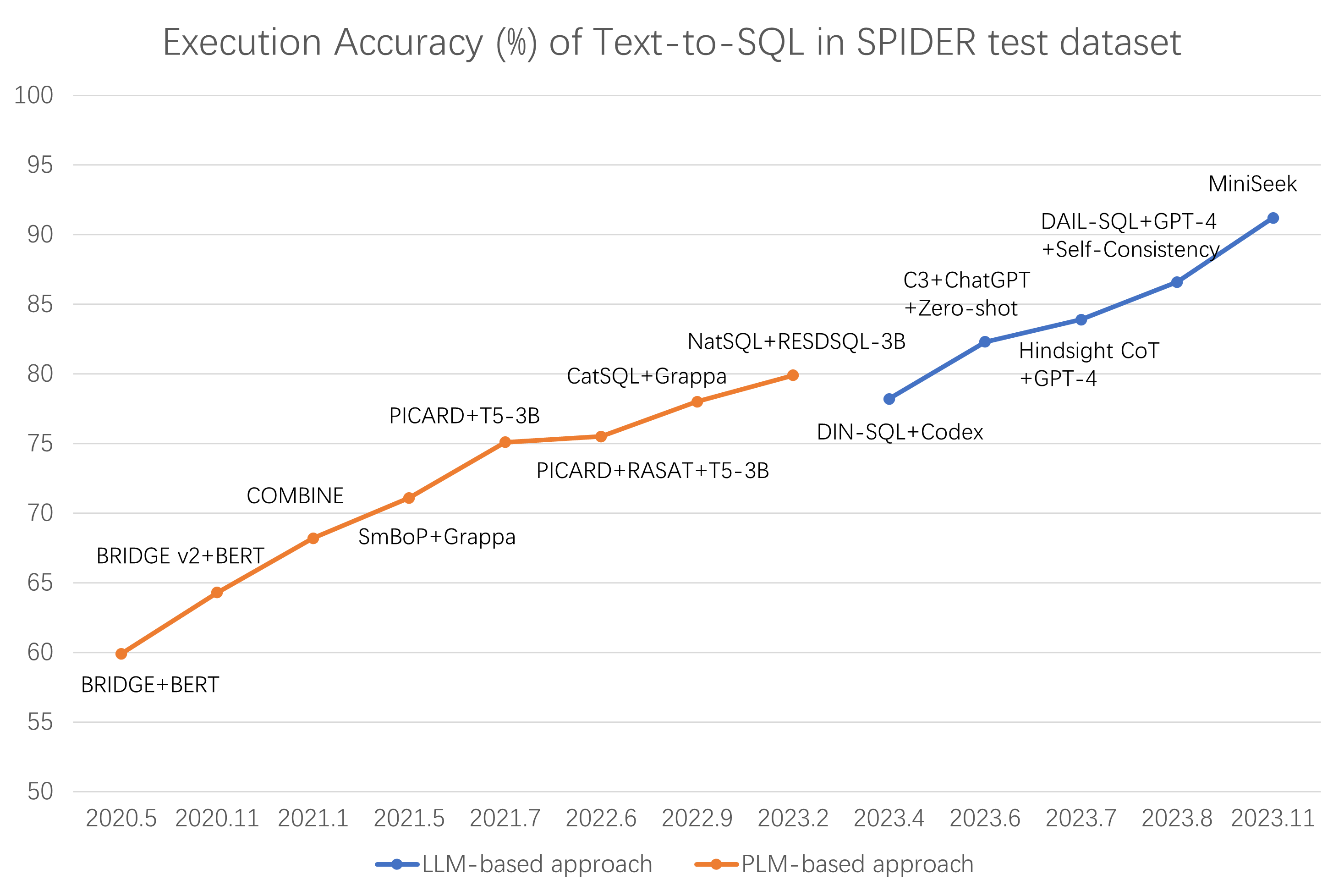}
    \caption{The evolution of Text-to-SQL approach over time}
    \label{fig:spider}
\end{figure}

There has been a sharp increase in the number of LLMs employed for text-to-SQL tasks in recent times. After conducting a comprehensive survey of recent papers, we identify several key reasons for this trend, which are summarized below:

\begin{itemize}
\item \textbf{Enhanced Performance}: Figure ~\ref{fig:spider} illustrates the progression of mainstreaming approaches within the Text-to-SQL domain, as indicated by the Execution Accuracy in the SPIDER test dataset\cite{yu2018spider}. As we can see, LLM-based methods have significantly improved the SOTA performance, demonstrating the remarkable capabilities of LLM-based methods.
The listed references\cite{lin-etal-2020-bridging, mellah2021combine, DBLP:journals/corr/abs-2010-12412, scholak2021picard, qi-etal-2022-rasat, li2023resdsql, pourreza2024din, dong2023c3, gao2023text, gan2021natural} support this observation.

\item \textbf{Generalization Ability and Adaptability}: As we mentioned above, LLMs have introduced a new paradigm, namely prompt engineering, which benefits from their ability to follow instructions\cite{wei2021finetuned}, making LLMs easily transferable to different settings without additional training. Furthermore, the in-context learning ability\cite{zhao2023survey} of LLMs further enhances their generalization and adaptability capabilities, as they can learn from examples provided to seamlessly fit into various settings.

\item \textbf{Future Improvements}: LLM-based methods hold a promising future for advancements. Since the global community is prioritizing the enhancement of LLMs, efforts and resources are concentrated towards supporting research on LLMs. This includes scaling up LLMs, creating new prompting methods, generating high-quality and extensive datasets, and fine-tuning LLMs across various tasks. The progress made by the LLM community will undoubtedly and continuously drive LLM-based text-to-SQL methods to new SOTA.

\end{itemize}

\section{Benchmark and Evaluation Metrics}

\begin{table}[]

\caption{Benchmarks basic information. The "\textbf{Supplement}" column contains a brief introduction to each dataset.}
\resizebox{.99\linewidth}{!}{
\begin{tabular}{lllll}
\toprule
Category & Dataset & \makecell[l]{Proposed\\time} & Statistics & Supplement\\
\hline
\multirow{17}{*}{\makecell[l]{single-\\turn}} 
& WikiSQL\cite{WikiSQL} & 2017 & \makecell[l]{80654 hand-annotated examples of questions and SQL queries\\ distributed across 24241 tables from Wikipedia } & From Wikipedia \\
\cline{2-5} 
& Spider 1.0\cite{yu2018spider} & 2018 & \makecell[l]{10,181 questions and 5,693 unique complex SQL queries on 200\\ databases with multiple tables covering 138 different domains} & \makecell[l]{A large-scale complex and cross-domain semantic parsing and text-to-SQL dataset annotated by 11 Yale\\ students} \\
\cline{2-5} 
& CSpider\cite{CSpider} & 2019 & \makecell[l]{9691 examples as Spider test set is not pubilcly available} & \makecell[l]{A Chinese large-scale complex and cross-domain semantic parsing and text-to-SQL dataset translated\\ from Spider } \\
\cline{2-5} 
& TABFACT\cite{chen2020tabfactlargescaledatasettablebased} & 2019 & \makecell[l]{117,854 manually annotated statements with regard to 16,573 \\Wikipedia tables} & \makecell[l]{TabFact is the first dataset to evaluate language inference on structured data, which involves mixed\\ reasoning skills in both symbolic and linguistic aspects} \\
\cline{2-5} 
& Spider-Realistic\cite{Spider-Realistic} & 2020 & \makecell[l]{508 examples on 19 databases}& \makecell[l]{Created based on the \textbf{dev} split of the Spider dataset and modified the original questions to remove the\\ explicit mention of column names} \\
\cline{2-5} 
& DuSQL\cite{DuSQL} & 2020 & \makecell[l]{containing 200 databases, 813 tables, and 23,797 question/SQL\\ pairs} & \makecell[l]{A larges-scale and pragmatic Chinese dataset for the cross-domain text-to-SQL task} \\
\cline{2-5} 
& Fiben\cite{sen:2020} & 2020 & \makecell[l]{300 NL questions, corresponding to 237 different complex SQL\\
queries on a database with 152 tables} & \makecell[l]{A dataset designed to evaluate the processing of complex business intelligence queries, often involving\\
nesting and aggregation, for NL2SQL systems} \\
\cline{2-5} 
& Spider-SYN\cite{ADVETA} & 2021 & \makecell[l]{Spider-Syn contains 7000 training
and 1034 development \\examples} & \makecell[l]{NL questions in Spider-Syn
are modified from Spider, by replacing their
schema-related words with manually\\ selected
synonyms that reflect real-world question paraphrases.} \\
\cline{2-5} 
& KaggleDBQA\cite{lee2021kaggledbqa} & 2021 & \makecell[l]{272 examples covering 8 DB} & \makecell[l]{A cross-domain evaluation dataset of real Web databases, with domain-specific data types, original formatting, \\and unrestricted questions} \\
\cline{2-5} 
& SPIDER-CG\cite{Spider-CG} & 2022 & \makecell[l]{45,599
examples, around six times the Spider dataset} & \makecell[l]{It split the sentences in the Spider into subsentences, annotate each sub-sentence with corresponding SQL \\clause and then compose these sub-sentences in different combinations to
test the ability of models to \\generalize compositionally} \\
\cline{2-5} 
& ADVETA\cite{ADVETA} & 2022 & \makecell[l]{depending on which datasets are used} & \makecell[l]{Based on some dataset and add adversarial table perturbation to fool Text-toSQL parsers} \\
\cline{2-5} 
& \makecell[l]{SCIENCEBENCH\\MARK\cite{zhang2023sciencebenchmarkcomplexrealworldbenchmark}} & 2023 & \makecell[l]{more than 6,000 NL/SQL-pairs} & \makecell[l]{A complex NL-to-SQL benchmark for three real-world, highly domain-specific databases: Research Policy \\Making, Astrophysics, and Cancer research}  \\
\cline{2-5} 
& DR.SPIDER\cite{chang2023dr.spider} & 2023 & \makecell[l]{ 15K perturbed examples} & \makecell[l]{Based on Spider and designs 17 perturbations on databases, natural language questions, and SQL queries to\\ measure robustness from various angles} \\
\cline{2-5} 
& BIRD\cite{li2024can} & 2023 & \makecell[l]{containing 12,751 text-toSQL pairs and 95 databases with a \\total size of 33.4 GB, spanning 37 professional
domains} & \makecell[l]{It highlights the new challenges of dirtyand noisy database values, external knowledge grounding between NL \\questions and database values, and SQL efficiency, particularly in the context of massive databases} \\
\cline{2-5} 
& Spider 2.0\cite{spider2} & 2024 & \makecell[l]{600 real-world text-to-SQL workflow problems} & \makecell[l]{The databases in Spider 2.0 are sourced from real data applications, often containing over 1,000 columns and\\ stored in cloud or local database systems such as BigQuery, Snowflake, or PostgreSQL. Solving problems in\\ Spider 2.0 frequently requires understanding and searching through database metadata, dialect documentation, \\and even project-level codebases.} \\
\cline{2-5} 
& BIRD-Critic 1.0\cite{li2024can} & 2025 & \makecell[l]{600 tasks for development and 200 held-out OOD tests} & \makecell[l]{It introduces a novel SQL evaluation framework designed to assess LLMs' ability to diagnose and resolve user \\ issues in real-world database environments.} \\
\midrule
\multirow{2}{*}{\makecell[l]{multi-\\turn}} 
& CoSQL\cite{yu2019cosql} & 2019 & \makecell[l]{30k+ turns plus 10k+ annotated SQL queries} & \makecell[l]{A corpus for building cross-domain Conversational text-to-SQL systems} \\
\cline{2-5}
& SParC\cite{SParC} & 2019 & \makecell[l]{4,298 coherent question sequences} & \makecell[l]{A dataset for cross-domain Semantic Parsing in Context} \\
\bottomrule
\end{tabular}
}
\label{benchmarks}
\end{table}

High-quality datasets are of crucial importance in the training and testing of Text-to-SQL tasks. In this section, we provide a summary of the mainstream benchmark datasets and evaluation metrics that are utilized in LLM-based text-to-SQL research papers. Additionally, we highlight key benchmarking studies that assess the NL2SQL performance and related capabilities of various large language models.

\subsection{Benchmark}

Upon comprehensive examination of papers on LLM-based text-to-SQL, we categorize text-to-SQL datasets into two categories: benchmarks prior to the rise of LLMs and benchmarks in the era of LLMs. We display the details of each dataset in Table~\ref{benchmarks}.

\textbf{Benchmarks prior to the rise of LLMs}. Classic benchmarks in this category, such as WikiSQL\cite{WikiSQL}, Spider 1.0\cite{yu2018spider}, and KaggleDBQA\cite{lee2021kaggledbqa}, are widely utilized both before and after the emergence of LLMs. They have made crucial contributions to the advancement of research in the Text-to-SQL domain. As of the time of our writing, Spider 1.0\cite{yu2018spider} remains the top choice for evaluating the performance of text-to-SQL methods. Based on these classic benchmarks, a variety of augmented benchmarks have been proposed. For instance, Spider-Realistic\cite{Spider-Realistic} creates scenarios where the explicit appearance of the database schema in the questions is eliminated. Spider-SYN\cite{ADVETA} replaces schema-related words in questions with manually selected synonyms that mirror real-world question paraphrases. SPIDER-CG\cite{Spider-CG} splits the sentences in Spider into sub-sentences, annotates each sub-sentence with the corresponding SQL clause, and then composes these sub-sentences in different combinations to test the models' ability to generalize compositionally. CSpider\cite{CSpider} translates Spider into Chinese to test the performance of Chinese text-to-SQL. ADVETA\cite{ADVETA} adds adversarial table perturbation to deceive Text-to-SQL parsers. There are also some other datasets prior to the rise of LLMs, such as Fiben\cite{sen:2020} and DuSQL\cite{DuSQL} and you can find brief introductions in Table~\ref{benchmarks}.

\textbf{Benchmarks in the era of LLMs} primarily denote benchmarks that emerged after the rise of LLMs. Due to the immense power of LLMs having propelled this field into a new phase, these benchmarks start to focus on more challenging aspects in real-world text-to-SQL tasks, including domain-specific knowledge, diverse perturbations, large and noisy databases, as well as SQL efficiency. Here, we've summed up five noteworthy benchmark datasets born after the widespread use of LLM.

\begin{itemize}
    \item \textit{SCIENCEBENCHMARK}\cite{zhang2023sciencebenchmarkcomplexrealworldbenchmark}. As pointed out in \cite{zhang2023sciencebenchmarkcomplexrealworldbenchmark}, previous datasets such as Spider are not representative of the difficulties encountered when creating an NL interface for a real-world database. Applying a system trained on those datasets to a new domain like astrophysics or cancer research yields poor results, making the adoption of such systems in real-life applications extremely implausible. Hence, SCIENCEBENCHMARK\cite{zhang2023sciencebenchmarkcomplexrealworldbenchmark} is proposed. It is the first text-to-SQL dataset developed in collaboration with SQL experts and researchers from the fields of research policy making, astrophysics, and cancer research.
    
    \item \textit{BIRD}\cite{li2024can}. Most datasets \cite{yu2018spider, ADVETA, spider-dk, Spider-Realistic, Spider-CG, ADVETA, CSpider, DuSQL, SParC, yu2019cosql, WikiSQL} focus on the database schema with few rows of database values, leaving the gap between academic research and real-world applications. To address this gap, \cite{li2024can} introduced the BIRD dataset. The BIRD dataset includes 12,751 text-to-SQL pairs and 95 databases, with a total size of 33.4 GB across 37 professional domains. It also highlights new challenges including dirty and noisy database values, external knowledge grounding between natural language (NL) questions and database values, and SQL efficiency, especially in the context of massive databases. Experimental results indicate that even the most effective text-to-SQL models, i.e. GPT-4\cite{achiam2023gpt}, only achieve 54.89\% in execution accuracy, which is still far from the human result of 92.96\%, proving that challenges still stand.
    
    \item \textit{Dr.Spider}\cite{chang2023dr.spider}. The work in \cite{chang2023dr.spider} pointed out that Text-to-SQL models are vulnerable to task-specific perturbations, but previous curated robustness test sets usually focus on individual phenomena. To bridge this gap, \cite{chang2023dr.spider} introduced the Dr.Spider dataset. Based on the Spider dataset\cite{yu2018spider}, Dr.Spider designs 17 perturbations on databases, natural language questions, and SQL queries to measure robustness from various angles. To collect more diversified natural question perturbations, \cite{chang2023dr.spider} leveraged large pre-trained language models (PLMs) to simulate human behaviors in creating natural questions in a few-shot manner. Experimental results indicate that even the most robust model encounters a 14.0\% overall performance drop and a 50.7\% performance drop on the most challenging perturbation.
    
    \item \textit{Spider 2.0}\cite{spider2}. As indicated in \cite{spider2}, real-world enterprise-level text-to-SQL workflows often involve complex cloud or local data across diverse database systems, multiple SQL queries in various dialects, and a wide range of operations from data transformation to analytics. However, previous datasets lack evaluation in these aspects. Therefore, Spider 2.0\cite{spider2} is proposed. It consists of 600 real-world text-to-SQL workflow problems and is sourced from real data applications such as BigQuery, Snowflake, or PostgreSQL.

    \item \textit{BIRD-Critic 1.0}\cite{li2024can}. Unlike the aforementioned benchmarks that primarily focus on SQL generation, BIRD-Critic 1.0 introduces a novel SQL evaluation framework designed to assess LLMs' ability to diagnose and resolve user issues in real-world database environments. Built upon authentic user-reported problems, this benchmark extends beyond simple SELECT queries to encompass a broader spectrum of SQL operations, thereby better reflecting practical application scenarios. BIRD-Critic 1.0 specifically targets reasoning challenges, as evidenced by the low 38.5\% pass rate achieved by even the most advanced models like o1-preview\cite{openai2024o1preview}.
    
\end{itemize}

\subsection{Evaluation Metrics}
We have summarized the following five metrics:

\textbf{Exact Set Match Accuracy (EM)} is determined by comparing the literal content of the generated SQL and the ground-truth SQL. Specifically, EM compares whether the SQL clauses of the generated SQL and the ground-truth SQL are consistent. However, there are multiple ways to express the same SQL problem, so the EM metric often underestimates the prediction accuracy of the model.

\textbf{Execution Accuracy (EX)} is concluded by comparing the execution results of the generated SQL and the ground-truth SQL. However, SQL with different logic may yield the same results when executed, so EX may also overestimate the prediction accuracy of the model.

\textbf{Test-suite Accuracy(TS)}\cite{zhong2020semantic} creates a small, concentrated database test suite from a large number of randomly generated databases. These databases have a high code coverage rate for accurate queries. During the evaluation process, it measures the annotation accuracy of the predicted queries in this test suite, effectively calculating the strict upper limit of semantic accuracy.

\textbf{Valid Efficiency Score(VES)}\cite{li2024can} includes SQL execution efficiency in the evaluation scope. The calculation formula is shown in equation \ref{ves}, where the hat symbol represents the predicted result, 1 is the indicator function, which is only 1 when the predicted SQL is equivalent to the correct SQL, and R is the square root of the ratio. Intuitively, the higher the correctness rate of the generated SQL, the higher the execution efficiency of the generated SQL, and the higher the VES value.
\begin{equation}
    VES = \frac{\sum^{N}_{n=1}\mathbf{1}(V_n, \hat{V}_n)\cdot \mathbf{R}(Y_n, \hat{Y}_n)}{N},
    \mathbf{R}(Y_n, \hat{Y}_n)=\sqrt{\frac{E(Y_n)}{E(\hat{Y}_n)}}
\label{ves}
\end{equation}

\textbf{ESM+}\cite{ascoli2024esm+} is proposed based on Exact Set Match Accuracy (EM) and applies new rules to LEFT JOIN, RIGHT JOIN, OUTER JOIN, INNER JOIN, JOIN, DISTINCT, LIMIT, IN, foreign keys, schema check and alias checks compared to EM. \cite{ascoli2024esm+} compare the performance of 9 LLM-based models using TS, EM, and ESM+ and the results indicate that ESM+ offers a substantial improvement by reducing the occurrences of both false positives and false negatives that commonly plague the earlier metrics(EM and TS).

\subsection{Benchmarking Studies}
Several benchmarking studies have been conducted to evaluate the performance and capabilities of LLMs in the Text-to-SQL task. The work in \cite{rajkumar2022evaluating} assess Codex\cite{chen2021evaluating} and GPT-3\cite{floridi2020gpt} models of different sizes using different prompt configurations on Text-to-SQL benchmarks. Their findings demonstrate that generative language models trained on code establish a strong baseline for Text-to-SQL tasks, and that prompt-based few-shot learning can perform competitively with fine-tuning-based approaches. The work in \cite{gao2023text} conduct a systematic and extensive comparison of prompt engineering strategies, encompassing aspects such as question representation, example selection, and example organization. Based on their findings, they propose DAIL-SQL~\cite{gao2023text}. The work in ~\cite{zhang2024benchmarking} design five distinct tasks, namely Text-to-SQL, SQL Debugging, SQL Optimization, Schema Linking, and SQL-to-Text, to comprehensively assess LLM capabilities across the entire Text-to-SQL workflow. Another work\cite{YangSLLMLZ24} evaluate schema generation methods, including DIN-SQL\cite{pourreza2024din}, C3\cite{dong2023c3}, RESDSQL\cite{li2023resdsql}, and their own schema linking approach. Unlike the aforementioned studies, which primarily focus on prompt engineering, DB-GPT-Hub\cite{zhou2024db} emphasizes fine-tuning of medium- to large-scale open-source LLMs and provides a modular, extensible codebase that supports a variety of mainstream models and experimental setups.

While these studies provide valuable insights, the rapid evolution of LLM-based Text-to-SQL methods and the frequent updates to models and techniques make benchmarking results quickly outdated. Therefore, we do not delve into specific benchmarking results here and instead refer interested readers to the original works.

\section{Prompt Engineering}

\begin{figure}
    \centering
    \includegraphics[width=0.95\linewidth]{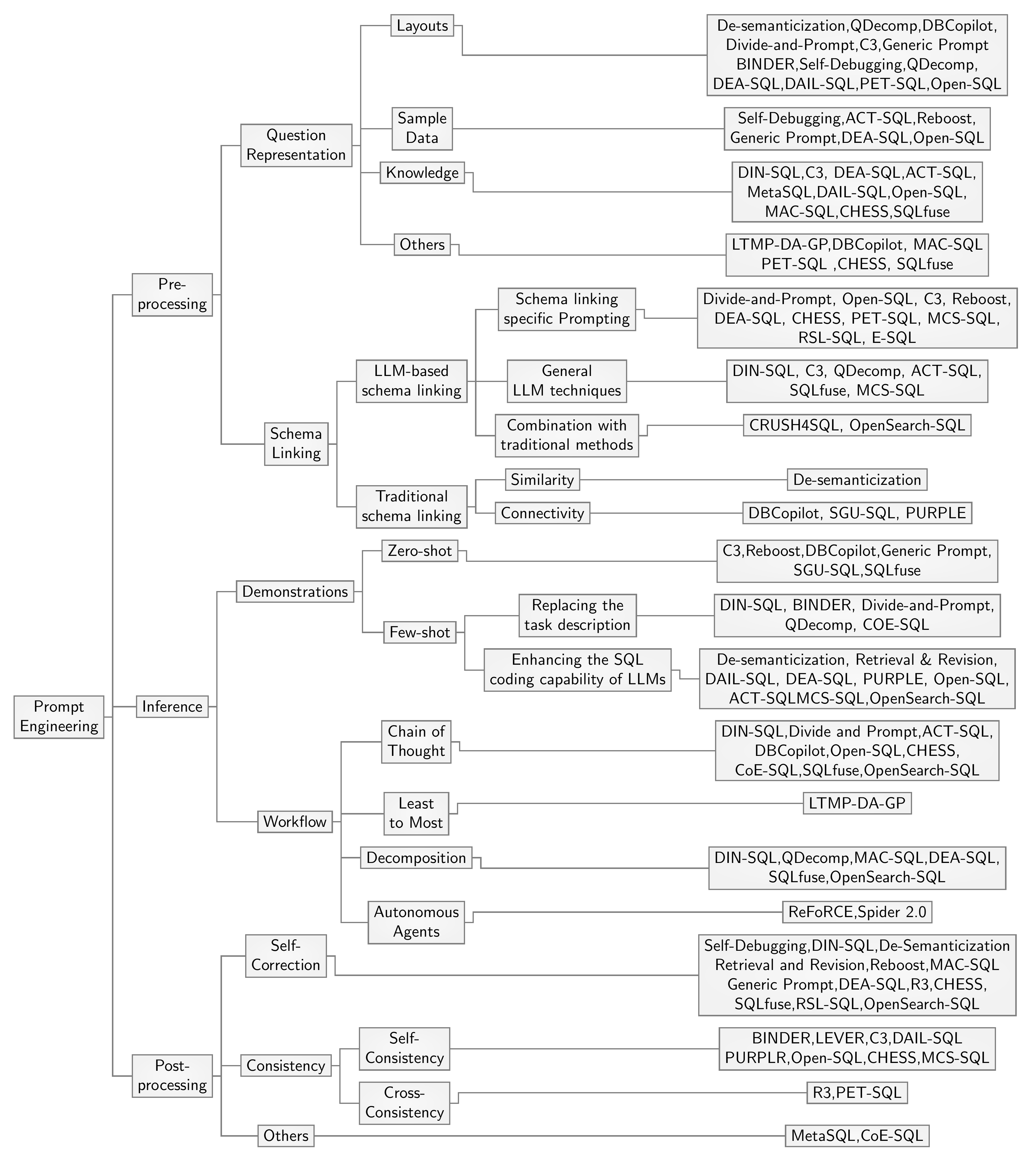}
    
    \caption{Taxomony of Prompt Engineering}
    \label{fig:prompt_taxomony}
\end{figure}

Prompt engineering, sometimes called in-context learning, is structuring the instructions that LLM can somehow understand. From the developer's perspective, it means customizing LLM's output on certain tasks by designing prompt words when interacting with LLM. Due to the autoregressive decoding properties\cite{fu2024break}, most LLMs predict the following text based on all currently visible preceding text which is also called context. We describe the autoregressive decoding as Equation~\ref{decoding}, as \textit{$y_t$} indicates the next token LLM will output, and \textbf{$x$} indicates prompt tokens given by users. The design of prompt words as context will affect the probability distribution of all upcoming tokens, affecting the final generation.

\begin{equation}
    y_t = \arg\max P(y_t|y_{1:t-1},\textbf{x})
\label{decoding}
\end{equation}

After the comprehensive investigation of related papers, we divide prompt engineering methods on text-to-SQL into the following three stages:
\begin{itemize}
    \item \textbf{Pre-processing}. In practical scenarios, experts engaged in text-to-SQL are often confused by the unclearness of question descriptions and the fuzziness of database schema. This situation also applies to LLMs and urges the pre-processing of the text-to-SQL problems. Our pre-processing part of the prompt engineering method will consist of two parts: first, the abstract representation of question descriptions; second, the selective linking of database schema.
        
    \item \textbf{Inference}. The so-called inference is defined as generating the corresponding SQL query when presented with the user's question and corresponding database schemas. This stage can be logically divided into two parts: workflow design and demonstration usage. Moving from pre-processing to SQL generation, most work will design their inference workflow either in a custom fashion or based on reasoning patterns, and decide whether to make use of demonstrations. We will start with the introduction of workflow design in text-to-SQL, including the well-known reasoning patterns like Chain-of-Thought\cite{wei2022chain}, Least-to-Most\cite{zhou2022least}, Decomposition, and autonomous agents. After that, we will introduce demonstration methods, including zero-shot methods and few-shot methods. In the few-shot section, we will emphasize the significance of demonstration style and selection.

    \item \textbf{Post-processing}. To enhance the performance and stability of LLM-based Text-to-SQL methods, it is optional to further optimize the generated results after inference. We call these operations as post-processing. Common post-processing methods in text-to-SQL include Self-Correction\cite{selfcorrection} and consistency methods (also known as Self-Consistency\cite{wang2022self} and Cross-Consistency\cite{r3, li2024pet}).
\end{itemize}

Figure~\ref{fig:prompt_taxomony} shows related works classified according to the above three parts, and the pipeline of LLM-based prompt engineering methods for text-to-SQL can be formulated as Equation~\ref{prompt_pipeline}. Detailed stages and insight are as follows.

\begin{equation}
    pred\_SQL = Post\_process(LLM(QuestionRepresentation,\ Demonstration,\ Reasoning))
    \label{prompt_pipeline}
\end{equation}

\subsection{Pre-processing}
At the beginning of text-to-SQL tasks, it's necessary to clearly and comprehensively describe all information needed to solve the problems in the prompt words. The information mainly contains the representation format of questions, database schema, and some task-related knowledge. In the pre-processing part, we will make an introduction to each information part above.

\subsubsection{Question Representation}
\begin{figure}
    \centering
    \includegraphics[width=0.75\linewidth]{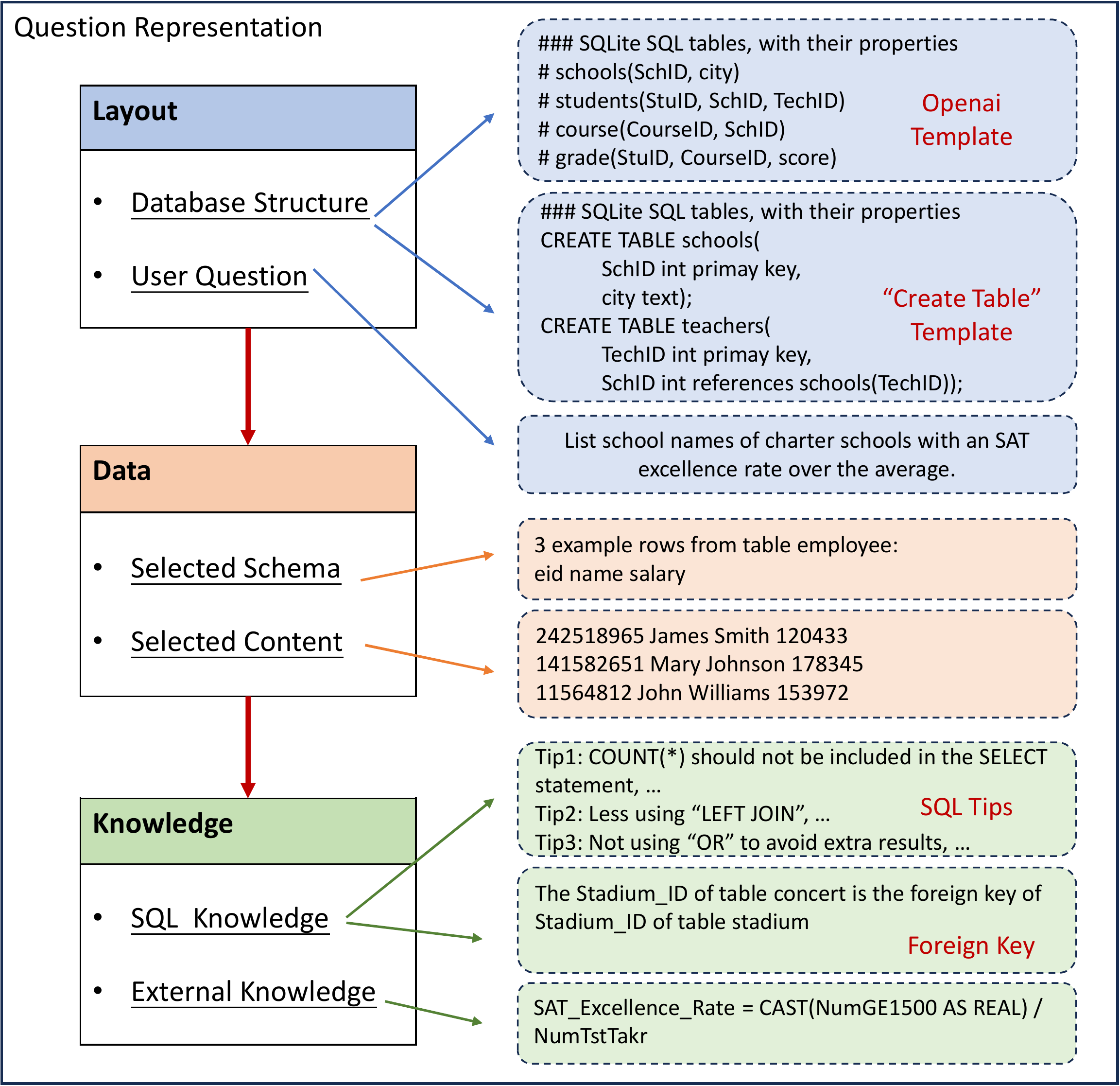}
    \caption{The framework of "Question Representation". It usually contains three parts sequentially. The \textit{Layout} part includes the question itself and the database structures. The \textit{Data} part includes sampled data from real database content. The \textit{Knowledge} part includes related evidence of some SQL-related priori and other knowledge from the external world. The detail in the figure is just for functional illustration.}
    \label{fig:QR}
\end{figure}

The "Question Representation" in text-to-SQL tasks includes two parts: the natural language problem statement and related databases' necessary information. Recent works\cite{aroraageneric,rajkumar2022evaluating,wang2023mac} explore commonly used question representations for questions and databases. The framework of "Question Representation" is shown in Figure~\ref{fig:QR}. We summarize some typical characteristics of these representations and describe them in the following.

\textbf{Layouts} One characteristic is the writing style of the question itself and the database structures, which we call the "layout" of the problem. Two common layouts are the OpenAI template layout and the "Create Table" layout.

\begin{itemize}
    \item \textit{Openai Template layout} Some work \cite{guo2023case,liu2023comprehensive,rajkumar2022evaluating,tai2023exploring,wang2023dbcopilot,liu2023divide,dong2023c3,aroraageneric,li2024pet} used the Openai prompt template of text-to-SQL task provided in the Codex official API documentation, with each line starting with the comment symbol of SQLite, followed by the form of Table (Column1, Column2,...). After completing the table description, the last row provides the SELECT word. The sample format is shown on the top-right part in Figure~\ref{fig:QR}.
    \item \textit{"Create Table" layout} Other work \cite{li2024can,nan2023enhancing,rajkumar2022evaluating,gao2023text,chen2023teaching,opensql} represented prompt words for database organization as CREATE TABLE creation statements, which is explained on the top-right part of Figure~\ref{fig:QR}. Explicitly including the data type of each column and possible primary and foreign key relationships in the statements will align the data format and perform multi-table connections when LLM generates SQL. Other works \cite{cheng2022binding,tai2023exploring,deasql} used CREATE TABLE statements and partial data to build prompt words, which is more in line with the real situation of data engineers writing SQL.
\end{itemize}

\textbf{Sample Data} Apart from the layouts, sample data from real database content is also beneficial for the question representation. The work in \cite{rajkumar2022evaluating} represented the prompt as a "SELECT * From Table LIMIT X" statement and the execution result of this SQL statement, where X is an integer and is often taken as 3 based on experience. Some work \cite{chen2023teaching,zhang2023act,sui2023reboost,aroraageneric,deasql,opensql} even directly put sample data into prompt words. The purpose is to enable LLM to understand the sample data in the database and conform to the format of the data itself when generating SQL. \cite{chang2023prompt} modified the form of Sample Data to list column-wise data, to explicitly enumerate categorical value.

Some works combine the layout design and sample data into their prompt to gain the advantage of both. Generic Prompt \cite{aroraageneric} concatenated the prompt words of Openai Template and Sample Data. LTMP-DA-GP\cite{arora2023adapt} added a random sampling of categorical columns and range information for numerical columns after CREATE TABLE statements. Besides, SQLfuse\cite{zhang2024sqlfuseenhancingtexttosqlperformance} added one-to-many relationship(a single record in one table is associated with multiple records in another) and enumerating value into the prompt; Some works \cite{kothyari2023crush4sql,pourreza2024din,wang2023mac,guo2023prompting,sun2023sql} also use the form of \textit{Table.Column} to list database schema.

\textbf{Knowledge} Besides layouts and sample data, it's also beneficial to provide related knowledge for LLM with general ability. The knowledge added to prompts can be viewed as a calibration of the current task description, clearing the way for the subsequent SQL generation. We divide knowledge into two categories: one is related to SQL, and the other is related to the problem itself and data.

\begin{itemize}
    \item \textit{SQL Knowledge} SQL-related knowledge mainly includes SQL keywords, SQL syntax, and common SQL writing habits. Adding SQL-related knowledge to the prompt for general LLM is like providing junior DBA with an experience handbook, which avoids semantic errors. C3\cite{dong2023c3} specifically calibrates the model's bias in SQL style and adds the following instructions in the prompt: using \textit{COUNT(*)} only in specific cases, avoiding using \textit{LEFT JOIN}, \textit{IN} and \textit{OR}, and using \textit{JOIN} and \textit{INTERSECT} instead, and recommending the use of \textit{DISTINCT} and \textit{LIMIT} keywords. DIN-SQL\cite{pourreza2024din} and DEA-SQL\cite{deasql} notice that some keywords like \textit{JOIN}, \textit{INTERSECT}, and \textit{IN} indicate the hardness of SQL queries, so they design different specifications and hints about the hardness level of the current question based on their judgment. Similarly, Meta-SQL\cite{metasql} designs three kinds of query metadata to be expressive enough to represent queries' high-level semantics. The types of metadata are operator tag, hardness value, and correctness indicator. To further refine the decision-making capabilities of the critic model, SQLfuse\cite{zhang2024sqlfuseenhancingtexttosqlperformance} supplements calibration hints that enumerate common errors. These hints serve to preemptively address potential missteps, enhancing the model’s ability to generate a superior SQL query. Many works \cite{chen2023teaching,gao2023text,zhang2023act,dong2023c3,nan2023enhancing,sun2023sql,deasql,li2024pet} also emphasize primary or foreign key descriptions, attempting to increase LLM's attention to relationships between tables.
    \item \textit{External Knowledge} There is also miscellaneous knowledge from the external environment that may be helpful for text-to-SQL tasks, for some jargon or domain-specific words are hard to understand without explanation. MAC-SQL\cite{wang2023mac} includes some additional informational evidence and database-related requirements before or after the question description. The goal is to clarify schema items and make the LLM clear about the meaning of special terms. Open-SQL\cite{opensql} utilizes additional descriptive information accompanying each query, as provided by the BIRD dataset\cite{li2024can}, that acts as a bridge between human comprehension and the database structure. CHESS\cite{chess} uses a context retrieval method to extract descriptions and abbreviations of database catalog, tables, and columns for better performance. SQLfuse\cite{zhang2024sqlfuseenhancingtexttosqlperformance} specifically designs an "SQL Critic" module to determine the optimal candidate SQL query. The module constructs an external SQL knowledge base from an array of intricate SQL statements and schema in GitHub to better extract knowledge in the external environment.
\end{itemize}

Following the experiments in \cite{dong2023c3,rajkumar2022evaluating,aroraageneric,tai2023exploring,gao2023text,zhang2023act,zhang2024sqlfuseenhancingtexttosqlperformance,deasql}, we will discuss the observations of performance with different question representations and share some takeaways.

\begin{itemize}
    \item \textit{Layout} In the layout perspective, C3\cite{dong2023c3} shows a significant drop when changing the layout from clear and structural mode to unstructured mode. When comparing the OpenAI template and the 'Create Table' template, DAIL-SQL\cite{gao2023text} claims that the former is better, while QDecomp\cite{tai2023exploring} argues that they are equally matched.
    \item \textit{Sample Data} The work in \cite{rajkumar2022evaluating,aroraageneric} provides performance statistics with each layout, indicating that sample data is pluggable and beneficial to other layouts. ACT-SQL\cite{zhang2023act} states that more samples are not always better.
    \item \textit{Primary and foreign keys} Many works show the importance of primary and foreign keys in the question representation. The ablation study in \cite{zhang2023act,deasql,li2024pet} claims that removing foreign keys will significantly decrease performance. SQLfuse\cite{zhang2024sqlfuseenhancingtexttosqlperformance} also puts primary keys as the second contributor to the accuracy gain. In DAIL-SQL\cite{gao2023text}, some LLMs exhibit gain while others see a drop after adding foreign keys.
    \item \textit{Knowledge} SQLfuse\cite{zhang2024sqlfuseenhancingtexttosqlperformance} and DEA-SQL\cite{deasql} show in their ablation study that the performance degrades significantly without extra information on SQL knowledge and external commonsense.
\end{itemize}

Based on the discussion above, we are ready to give some takeaways on the choice of question representation methods.

\begin{framed}
\textbf{Key takeaways}
\begin{itemize}
    \item Structural layouts (like the Openai template and the 'Create Table' template) are better than the unstructured layouts, while different structural layouts are equally matched.
    \item Sample data is effective, pluggable, and worth consideration when the context length is sufficient.
    \item Primary and foreign keys have significant effects and are critical and applicable in complex scenarios.
    \item The knowledge part is important for LLM to understand some SQL evidence and hidden concepts in the question and align with commonsense.
\end{itemize}
\end{framed}

\begin{figure}
    \centering
    \includegraphics[width=\linewidth]{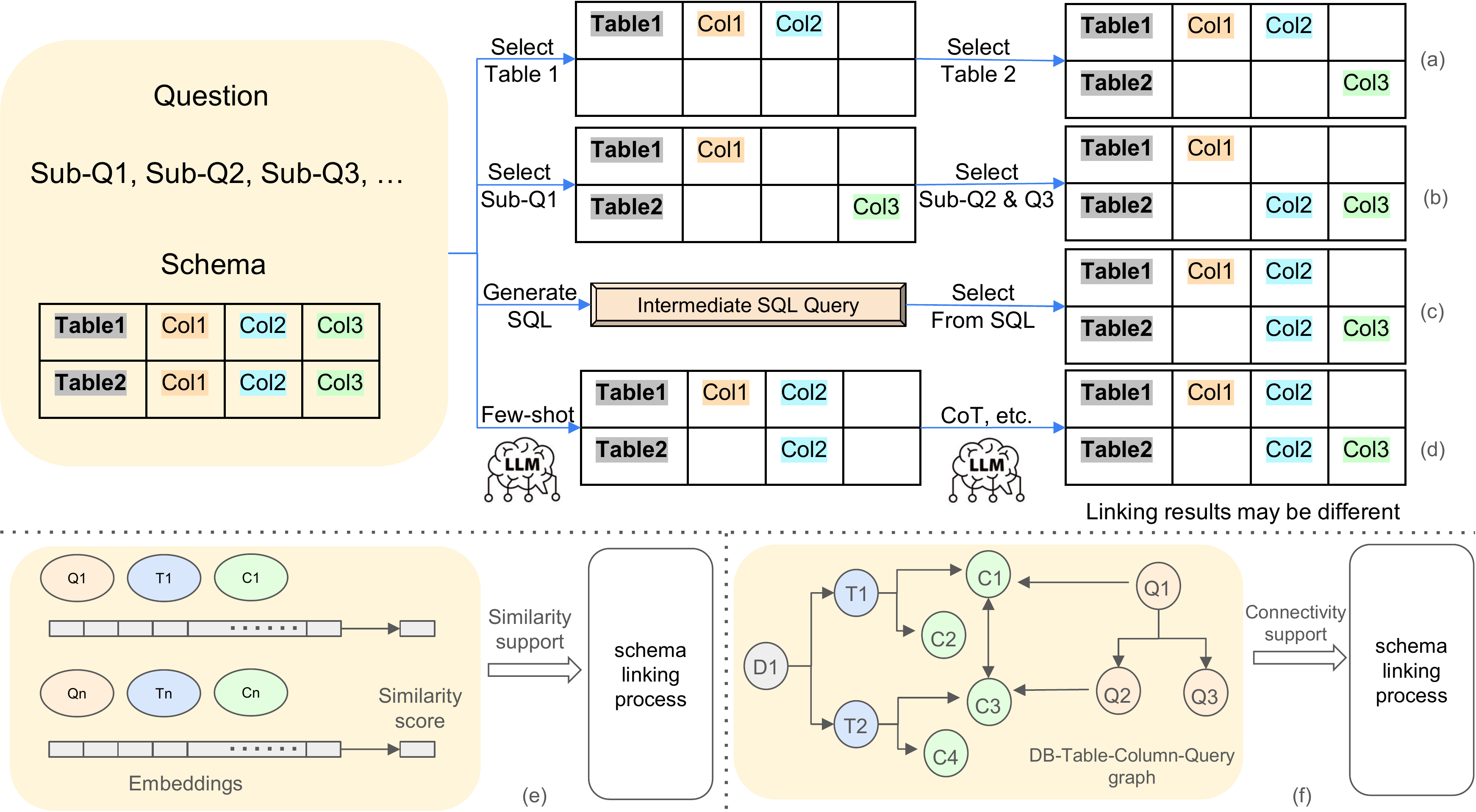}
    \caption{An overview of schema linking methods used in LLM-based text-to-SQL papers. (a), (b), (c), and (d) are LLM-based schema linking methods, while (e) and (f) are traditional schema linking methods. (a) and (b) correspond to prompting LLMs in specific steps designed for schema linking. (c) corresponds to using SQL to guide schema linking. (d) corresponds to enhancing LLM-based schema linking performance by utilizing general LLM techniques. (e) and (f) are similarity methods and connectivity methods, respectively.}
    \label{fig:schema_linking}
\end{figure}

\subsubsection{Schema Linking}
There are often situations where a problem involves multiple columns in multiple tables, but each table only uses specific columns.  Redundant and unrelated schema items probably distract LLM from identifying the right items, calling for schema linking. Schema linking is the subtask within the text-to-SQL process to specify the tables and columns in the database that correspond to the phrases in the given query. Tables and columns can be specified together or separately \cite{Distillery}. Schema linking is crucial in LLM-based text-to-SQL pipelines for two reasons. First, it brings about shorter token lengths. On one hand, for a large database, it is impractical to prompt all the table descriptions into the LLM. On the other hand, by reducing the length of the table schema, we can potentially enhance the attention and effectiveness of the LLMs. Second, research has shown that a significant number of failures in LLM-based text-to-SQL stem from the inability to correctly identify column names, table names, or entities mentioned in questions. By utilizing schema linking techniques, we can improve the performance and even facilitate cross-domain generalization and synthesis of complex queries.

After conducting a comprehensive investigation of schema linking techniques employed in recent LLM-based text-to-SQL papers, we have categorized them into two distinct categories, namely LLM-based schema linking methods and traditional schema linking methods. Figure~\ref{fig:schema_linking} shows an overview of our surveyed schema linking methods.

\textbf{LLM-based schema linking methods} refers to methods that utilize LLMs to perform the schema linking task. There are three main approaches:
\begin{itemize}
    \item \textit{Prompting LLMs in specific steps designed for schema linking.} The simplest way is to directly prompt LLMs to do schema linking, as Divide-and-Prompt\cite{liu2023divide} and Open-SQL\cite{opensql} do. To further improve the performance in this way, some studies \cite{dong2023c3, sui2023reboost, deasql, li2024pet, chess} designed more complex paths which prompt LLMs in specific steps. C3\cite{dong2023c3} and MCS-SQL\cite{mcssql} break down schema linking into two steps. They first instruct the LLM to recall tables and then retrieve the columns within the candidate tables. Reboost\cite{sui2023reboost} adopts the same approach as C3\cite{dong2023c3} but enriches table and column selection by including query descriptions and column explanations. DEA-SQL\cite{deasql} explored first identifying elements in a query and then utilizing them to filter the schema. CHESS\cite{chess}, on the other hand, follows a three-step paradigm of "column filtering, table selection, and final column filtering". PET-SQL\cite{li2024pet} emphasizes that LLMs excel in writing SQL compared to the schema linking sub-task, proposing an initial step of composing corresponding SQL queries and then extracting tables and columns from these queries. RSL-SQL \cite{rslsql} introduces a bidirectional schema linking approach to consider both full schema and SQL generated by the LLM, combining the methods and advantages of MCS-SQL\cite{mcssql} and PET-SQL \cite{li2024pet}. Also equipped with the guidance of the pre-generated SQL candidate, E-SQL \cite{esql} chooses to enrich the natural language query by directly incorporating relevant database items and conditions into the question and SQL generation plan.
    
    \item \textit{Enhancing LLM-based schema linking performance by utilizing general LLM techniques.} As LLM-based schema linking methods are driven by LLMs, several studies\cite{pourreza2024din, tai2023exploring, dong2023c3, zhang2023act, zhang2024sqlfuseenhancingtexttosqlperformance} explored enhancing schema linking performance by leveraging general LLM techniques, such as few-shot learning\cite{brown2020language}, chain-of-thought reasoning\cite{wei2022chain}, self-consistency voting\cite{wang2022self}, and fintuning. DIN-SQL\cite{pourreza2024din} randomly selects some examples to guide schema linking and utilizes "Let's think step by step" to further improve performance. C3\cite{dong2023c3} employs self-consistency\cite{wang2022self} to enhance performance stability. MCS-SQL \cite{mcssql} makes different prompts to do schema linking and uses a union operation to include the selected schema items. QDecomp\cite{tai2023exploring} and ACT-SQL\cite{zhang2023act} also use few-shot learning to guide schema linking but place more emphasis on example construction. Concretely, given a sub-question and its corresponding SQL, they annotate all table-column pairs mentioned in the SQL as ground truth. In addition, ACT-SQL\cite{zhang2023act} also employs an embedding model to identify relationships between phrases and schemas, which are then used to construct examples in a chain-of-thought style. Besides the above prompting methods, SQLfuse\cite{zhang2024sqlfuseenhancingtexttosqlperformance} employs finetuning to enhance the LLM's schema linking performance.
    
    \item \textit{Integrating LLMs into traditional schema linking methods.} Besides the above methods, there is another option available, which involves integrating LLMs with traditional schema linking approaches. One such example is CRUSH4SQL\cite{kothyari2023crush4sql}, which takes advantage of the hallucination capability of LLMs. It first generates a DB schema based on the given query through hallucination, and then employs similarity-based retrieval to select related schemas using the generated schema as a reference. OpenSearch-SQL \cite{OpenSearch-SQL} utilizes vector retrieval after employing an LLM to select schema items to control the similarity between the original query and the selected items.
    
\end{itemize}

\textbf{Traditional schema linking methods} refers to schema linking techniques without relying on LLMs. Our thorough analysis reveals that LLM-based text-to-SQL research papers utilize two main streams of these traditional schema linking methods:
\begin{itemize}
    \item \textit{Similarity-based Retrieval Methods.} To retrieve relevant schemas, the basic idea is to determine the similarity between the query and the schema information and retrieve the most similar schemas. A common implementation employs pre-trained language models such as BERT\cite{devlin2018bert} and RoBERTa\cite{liu2019roberta} to encode both queries and schema elements into embedding vectors for similarity comparison. One such example is De-semanticization\cite{guo2023prompting}, which calculates similarity by identifying direct matches between each token in the question and each item in the schema, as well as identifying correspondences between question words and specific database values.
    
    \item \textit{Connectivity methods.} Besides similarity, connectivity is another essential consideration, as the tables and columns used should exhibit a relationship or connection between them. Some recent works, like DBCopilot\cite{wang2023dbcopilot}, PURPLE\cite{ren2024purplemakinglargelanguage}, and SGU-SQL\cite{zhang2024structureguidedlargelanguage}, have explored this concept by utilizing graph-based approaches. DBCopilot\cite{wang2023dbcopilot} first constructs a graph to represent the underlying schema structure of all databases and their tables, and then trains a Seq2Seq model as a router routing over massive databases to get a set of schemas. PURPLE\cite{ren2024purplemakinglargelanguage} incorporates graph creation based on foreign-primary key connections to enhance the inter connectivity of the retrieved relevant schemas. SGU-SQL\cite{zhang2024structureguidedlargelanguage} constructs a query-schema graph by combining the query structure and database structure based on the topic concepts in the query, predefined relations in the query, database schema as well as the table/column names present within the schema. Then it trains a model to bind the query nodes with corresponding schema nodes.
\end{itemize}

With the development of LLMs and the enlargement of the context window, more and more schema items can be put into the prompt of text-to-SQL tasks. The experiments in Distillery \cite{Distillery} show that as the model’s SQL generation capability improves, its sensitivity to the presence of irrelevant columns as contextual information for generation decreases.
The importance of schema linking diminishes as costs decrease, context windows widen, and generation capabilities improve. However, schema linking is still useful in most practical cases because the entire schema in real-world data-warehousing scenarios often exceeds the context window.

\begin{framed}
\textbf{Key takeaways}
\begin{itemize}
    \item Most works prefer Chain-of-Thought or decomposition reasoning as the foundational workflow.
    \item The mainstream LLM-based schema linking methods comprise prompting LLMs in specific steps and enhancing performance by utilizing general LLM techniques (such as few-shot learning, chain-of-thought reasoning, self-consistency voting, and finetuning).
    \item Both similarity and connectivity are worthy considerations.
\end{itemize}
\end{framed}

\subsection{Inference}
Given the question and schema in a certain form, the next step is to generate potential answers to the problem. Considering the complexity and high requirement for accuracy of text-to-SQL tasks, it's difficult to get satisfactory answers by letting LLM directly generate SQL responses to the problem. We investigate related works comprehensively and find two techniques that can help generate the right and high-quality SQLs: workflow and demonstrations.

\subsubsection{Workflow}
The simplest workflow is to generate SQL directly from the constructed question and schema by only one interaction with LLMs, which somehow overestimates the ability of general LLMs in professional fields. Just like people tend to break down a complex task into several simple sub-tasks or steps, prompt-engineering-based methodologies usually design specific inference workflows for generating responses to queries using LLMs. Workflows in text-to-SQL tasks can be categorized based on distinct reasoning patterns. The following paragraphs will introduce typical workflows in our scenario and compare their strengths in different contexts.

\textbf{Chain-of-Thought(CoT)} The most famous reasoning style, Chain-of-Thought (CoT) \cite{wei2022chain}, involves a series of intermediate reasoning steps and typically begins with the phrase "Let's think step by step" to elicit chain thinking. The reasoning steps are like clause by clause and keyword by keyword, and outputs the steps in a single pass.  DIN-SQL\cite{pourreza2024din}  employs human-designed steps in CoT for its complex class of questions. For nested complex questions, the CoT steps usually encompass several sub-questions of the original question, because they match the corresponding keywords or clauses in the SQL. Divide-and-Prompt\cite{liu2023divide} generates SQL in a clause-by-clause manner, a variant of the CoT pattern. CoE-SQL\cite{coe} also proposes a variant of CoT, Chain-of-Edition, which describes 14 edit rules of SQL statements: editing "SELECT" items, editing "WHERE" logical operator, etc. Furthermore, ACT-SQL\cite{zhang2023act} proposes Auto-CoT to generate CoT exemplars automatically, addressing the high cost of manually labeling CoT prompts. Open-SQL\cite{opensql} designs the CoT template that employs a skeleton-based query framework as the intermediate representation. 

\textbf{Least-to-Most} The Least-to-Most\cite{zhou2022least} reasoning style is another widely utilized approach in the design of text-to-SQL workflows. Different from Chain-of-Thought that usually focuses on the SQL part, Least-to-Most follows the principle that sub-problems can be reduced to the original problem syntactically and semantically. LTMP-DA-GP\cite{arora2023adapt} exemplifies this category. The paper adopts Least-to-Most to decompose the natural language query, map NatSQL\cite{gan2021natural} to the decomposition, and generate SQL from NatSQL.

\textbf{Decomposition} Apart from the above reasoning methods, a straightforward yet effective workflow exists that primarily decomposes the generation task into customized interactions with LLM, employing various techniques to address challenges at each stage\cite{khot2022decomposed}. QDecomp\cite{tai2023exploring} first claims this category by proposing the decomposition prompt method. The decomposition of the SQL generation task can be parallel or sequential. Starting from the error analysis of SQL statements, DIN-SQL \cite{pourreza2024din} classifies SQLs into three levels of complexity: \textit{Easy}, \textit{Nested Complex}, and \textit{Non-Nested Complex}. This is the parallel decomposition for the classification of the task. MAC-SQL\cite{wang2023mac} introduces a framework comprising a selector, a decomposer, and a refiner. The decomposer agent generates a series of intermediate steps (i.e. sub-questions and SQLs) before predicting the final SQL. This is the sequential decomposition. DEA-SQL\cite{deasql} follows the global decomposition steps of Information determination, Classification \& Hint, SQL generation, Self-correction, and Active learning. OpenSearch-SQL \cite{OpenSearch-SQL} also involves five modules: Preprocessing, Extraction, Generation, Refinement, and Alignment. SQLfuse\cite{zhang2024sqlfuseenhancingtexttosqlperformance} organizes its SQL generation prompt in the decomposition pattern to incorporate prior schema information and iteratively execute SQL checking and error correction within a single prompt. R3\cite{r3} proposes a consensus-based decomposition system for text-to-SQL tasks. The system consists of an SQL writer and several reviewers of different roles. After several rounds of “negotiation” between the SQL writer and reviewers, there will be a consensus, and the final answer will be decided.

\textbf{Autonomous Agents} Inspired by ReAct framework \cite{yao2022react}, autonomous agents can be adapted to the Text-to-SQL domain. Different from "multi-agent" frameworks claimed by much work, the autonomous-agent workflow treats the LLM as the brain and lets the LLM control the interaction with the environment, humans, and other LLM-based agents. An autonomous-agent framework usually features multi-turn interactions, extended inference time, long-term memory, and conditional termination. Spider 2.0 \cite{spider2} proposes Spider-Agent, which is primarily focused on database-related coding tasks and projects. Spider-Agent allows for multi-turn interactions with the database via command-line interfaces, and automatically terminates if it outputs the same result three times in a row or a timeout is triggered. REFORCE \cite{ReFoRCE} also proposes an agent-based framework with a self-refinement workflow and more domain-specific techniques like table compression, format restriction, and task description. The autonomous agents will not terminate until self-consistency, maximum iterations, or empty results.

Chain-of-Thought, Least-to-Most, Decomposition, and autonomous agents are all useful in the performance of text-to-SQL tasks. At the same time, they have their advantages in different contexts \cite{khot2022decomposed}. Chain-of-Thought generates SQL queries in a step-by-step manner, like clause by clause and keyword by keyword, and outputs the steps in a single pass. Least-to-Most mainly derives the sub-questions for problem reduction. Both Chain-of-Thought and Least-to-Most imitate humans' thinking processes and output understandable answers. Decomposition makes use of various techniques to solve the whole task, and it is suitable when users need additional preparations before constructing SQL queries or post-processing after the SQL query is generated. The preparations include schema linking, example selection, difficulty level judgment \cite{pourreza2024din}, SQL refinement \cite{deasql,wang2023mac}, and any other customized operations. Autonomous agents are an emerging method and show great potential in real-world tasks \cite{spider2} by letting LLMs think and interact with each other multiple times with long-term memory.

\begin{framed}
\textbf{Key takeaways}
\begin{itemize}
    \item Most works prefer Chain-of-Thought or decomposition reasoning as the foundational workflow.
    \item It's recommended to design customized variants of CoT compared with the original CoT.
    \item Decomposition can benefit SQL generation in a sequential or parallel way.
    \item Alternative patterns are open for exploration.
\end{itemize}
\end{framed}

\subsubsection{Demonstrations}
In the workflow of LLM-based text-to-SQL methods, demonstrations are frequently employed to enhance the performance of SQL generation by incorporating several demonstrations. Depending on whether demonstrations are appended, these methods can be classified into two categories: zero-shot methods and few-shot methods. Examples for zero-shot LLM-based text-to-SQL methods include C3\cite{dong2023c3}, ReBoost\cite{sui2023reboost}, DBCopilot\cite{wang2023dbcopilot}, Generic\cite{aroraageneric}, SGU-SQL\cite{zhang2024structureguidedlargelanguage} and SQLfuse\cite{zhang2024sqlfuseenhancingtexttosqlperformance}. Although zero-shot methods have the advantage of saving the tokens of LLMs, text-to-SQL is a complex task that may be relatively less familiar from the model's perspective. Merely modifying the instructions cannot effectively solve this complex task. In contrast to zero-shot, few-shot has the ability to learn the pattern of tasks from examples and does not solely depend on the pre-trained knowledge of the model. This characteristic enhances the performance and adaptability of LLMs. After conducting a comprehensive investigation of demonstrations employed in recent LLM-based text-to-SQL papers, we have categorized the functions of demonstrations into two distinct categories, including:
\begin{itemize}
    \item \textit{Replacing the task description}. As mentioned earlier, to handle complex text-to-SQL tasks, studies explore diverse workflows that comprise specific steps or involve newly defined subtasks that did not previously exist. For example, when generating SQLs for non-nested complex questions\cite{pourreza2024din}, DIN-SQL\cite{pourreza2024din} follows a workflow in the style of Chain of Thought with fine-designed steps. It stipulates that first, the NatSQL\cite{gan2021natural} corresponding to the problem is generated as the intermediate representation. Subsequently, the final result is generated based on this NatSQL\cite{gan2021natural}. Another example is BINDER\cite{cheng2022binding}, which augments programming languages (such as SQL) with language model API call functions and employs a large language model to translate natural language queries into extended SQL language. However, both DIN-SQL's\cite{pourreza2024din} fine-designed steps and BINDER's\cite{cheng2022binding} translation are unfamiliar from the model's perspective and are not easily described by human language prompts. In this situation, both of them make use of the in-context learning capability of LLMs and provide several examples. This significantly simplifies the instruction design in prompts. This idea is also employed by Divide-and-Prompt\cite{liu2023divide}, QDecomp\cite{tai2023exploring} and COE-SQL\cite{coe}.
    
    \item \textit{Enhancing the SQL coding capability of LLMs}. Appropriate demonstrations can greatly enhance the performance of 
    LLMs\cite{brown2020language} and experiments have found that LLMs are highly sensitive to sample selection, and choosing inappropriate samples may even produce negative effects. To maximize performance, studies\cite{guo2023prompting, retrievalandrevision, gao2023text, zhang2023act, deasql, ren2024purplemakinglargelanguage, opensql, mcssql} explore how to select appropriate demonstrations. The simplest way is to retrieve examples whose questions have similar semantic meanings. However, questions with different database schemes can vary significantly, even if their underlying intentions are similar and the corresponding SQL queries show similarities \cite{guo2023prompting}. 
    To bridge this gap, studies such as De-semanticization\cite{guo2023prompting}, Retrieval \& Revision\cite{retrievalandrevision}, DAIL-SQL\cite{gao2023text}, DEA-SQL\cite{deasql}, and PURPLE\cite{ren2024purplemakinglargelanguage} first mask domain-specific words in the original question to obtain the skeletons of the query. Then, they retrieve examples whose question skeletons have similar semantic meanings. Among them, PURPLE\cite{ren2024purplemakinglargelanguage} further designs four levels of SQL skeleton abstraction and focuses on more coarse-grained retrieval. Some studies\cite{retrievalandrevision, gao2023text, opensql, mcssql} try to use more information for retrieving examples. Retrieval \& Revision\cite{retrievalandrevision} simplifies natural language questions by prompting the LLM with instructions and retrieves examples utilizing both the original question and the simplified question, which avoids the frustration of unusual questioning styles and enhances the syntax and wording variety in the repository. DAIL-SQL\cite{gao2023text} retrieves examples utilizing both natural language questions and corresponding SQL queries, while Open-SQL\cite{opensql} retrieves examples utilizing natural language questions, database schema, and corresponding SQL queries. OpenSearch-SQL \cite{OpenSearch-SQL} proposes the mask question similarity to select similar queries. MCS-SQL \cite{mcssql} uses two distinct selection strategies to leverage both question similarity and masked question similarity. Besides retrieving similar examples, diverse examples may also be helpful. One example is ACT-SQL\cite{zhang2023act}. It utilizes randomly selected examples as well as examples similar to the current question. In addition to example selection, token-cost is another consideration. DAIL-SQL\cite{gao2023text} considers the trade-off between accuracy and token-cost and provides three example styles, namely the combination of the query, schemas and corresponding SQL, the combination of the query and corresponding SQL, and only SQL. DAIL-SQL\cite{gao2023text} chooses the combination of the query and corresponding SQL to reduce the token length of examples.
\end{itemize}

\begin{framed}
\textbf{Key takeaways}
\begin{itemize}
    \item One straightforward way of task description in a prompt is to utilize demonstrations. The more complex the workflow, the greater this method's advantage.

    \item Question skeletons can capture the question intention more effectively than the original questions. 

    \item A trade-off between accuracy and token-cost needs to be taken into account. 
\end{itemize}
\end{framed}

\subsection{Post-processing}
To further improve the performance and stability of LLM-based Text-to-SQL methods, studies apply post-processing to the generated SQLs. After a comprehensive investigation, we summarize these methods into two categories: self-correction and consistency.

\subsubsection{Self-Correction}
After LLM generates an answer, the Self-Correction method uses rules under specific questions and tasks to have LLM check the correctness of the answer; In the text-to-SQL scenario, the Self-Correction method can use SQL-related rules for checking, and can also provide LLM with the results or error logs generated by running SQL statements for inspection. \cite{aroraageneric} noticed the subtle difference in extra spaces in table values and let LLM recheck them. RSL-SQL \cite{rslsql} and OpenSearch-SQL \cite{OpenSearch-SQL} repeat the SQL generation until the results are not empty or the maximum dialogue rounds are reached. The approach of \cite{guo2023case,guo2023prompting,retrievalandrevision} is to change the small number of schema in the original prompt words to a full number of schema and generate SQL again if the SQL generated by LLM cannot run. In \cite{pourreza2024din,wang2023mac,sui2023reboost,retrievalandrevision,chess}, the incorrect SQL and the information indicating the execution error will be used as prompts for LLM to regenerate the SQL. \cite{deasql} analyzed several significant error points of field matching and SQL syntax, then designed specific prompts to correct them. \cite{chen2023teaching} designed unit tests, code explanation, and execution results for the model to refine its response. \cite{zhang2024sqlfuseenhancingtexttosqlperformance} proposed the SQL Critic module and adopted a few-shot in-context learning strategy that leverages examples from an external SQL knowledge base, enriched with hindsight feedback.

\subsubsection{Consistency Method}
The Self-Consistency method \cite{wang2022self} mainly adopts a majority voting strategy, allowing the same LLM to generate multiple answers to the same question by setting a certain temperature. Then LLM selects the answer that appears the most frequently as the final answer. In some work \cite{cheng2022binding,dong2023c3,sun2023sql,gao2023text,opensql,chess,OpenSearch-SQL}, the Self-Consistency method was directly used to perform a major vote on the SQL generated by its LLM, achieving good performance improvement. \cite{ni2023leverlearningverifylanguagetocode} proposed to rerank multiple answers and trained a verifier to check the generated code. PURPLE\cite{ren2024purplemakinglargelanguage} utilized the majority vote based on execution results of generated SQLs. MCS-SQL \cite{mcssql} utilizes multiple prompts to generate a certain number of SQLs for each prompt, and then selects the most accurate query among the candidates based on the confidence scores and execution time.

Unlike Self-Consistency, the Cross-Consistency method uses several different LLMs or agents to generate answers respectively or check the validity of SQLs. PET-SQL \cite{li2024pet} proposed this method to instruct several LLMs under lower temperatures to generate SQLs and then make votes across the SQLs’ executed results. To leverage the advantages of respective strengths, \cite{chess} used Self-Correction and Self-Consistency sequentially in its post-processing stage. R3 \cite{r3} adopted a multi-agent framework to provide suggestions on SQL from agents with different expertise in a loop way, like the combination of Cross-Consistency and Self-Correction.

\subsubsection{Others}
DEA-SQL\cite{deasql} discovers that the model is more prone to errors for certain problem types (e.g., extremum problems), and then utilizes active learning\cite{settles2009activelearning} to represent the three fixed error cases and to identify whether the generated SQL should be modified, which better meets the requirements. OpenSearch-SQL \cite{OpenSearch-SQL} focuses on the consistency alignment in the process of SQL generation and adds an alignment module to check each module in the framework aligns with specific requirements.

\begin{framed}
\textbf{Key takeaways}
\begin{itemize}
    \item In the text-to-SQL tasks, most works equipped with Self-Correction focus on the 'refine' parts like some handmade schema-modify rules and execution logs, while the 'critic' parts like code explanation and answer judgment have more room for exploration.
    \item Self-Consistency has the advantages of adaptability, convenience, and good performance. However, it needs more interactions with LLM and costs more.
    \item Cross-Consistency lets multiple LLMs participate, reducing the prejudice drawback of a single LLM.
    \item It's promising to use both Self-Correction and consistency methods in the post-processing stage.
\end{itemize}
\end{framed}
\section{Fine tuning}
LLMs have demonstrated remarkable performance on Text-to-SQL tasks through prompting methods such as Retrieval-Augmented Generation (RAG), In-Context Learning (ICL), and Chain-of-Thought (CoT)\cite{wei2022chain}. Nevertheless, these methods heavily rely on powerful yet closed-source LLMs like GPT-4\cite{achiam2023gpt}, which may cause privacy concerns\cite{Db-gpt, codeS} since data needs to be transmitted to closed-source model providers. Therefore, we should also pay attention to the application of open-source LLMs in Text-to-SQL tasks. Compared to large closed-source LLMs like GPT-4\cite{achiam2023gpt}, open-source LLMs suffer from limited reasoning and instruction-following capabilities. This is due to several factors, including smaller model sizes, a pre-training corpus that is both smaller in quantity and less qualified in quality, as well as less extensive pretraining efforts. Additionally, SQL-related content typically accounts for only a minuscule fraction of their entire pre-training corpus\cite{codeS}, potentially resulting in a subpar performance in text-to-SQL tasks. Therefore, the mainstream of application of open-source LLMs in Text-to-SQL tasks is finetuning.

In the context of LLM-based text-to-SQL, finetuning involves taking a base LLM, for instance, code-llama\cite{Code-llama}, and conducting further training on text-to-SQL related tasks, typically the SQL generation task, by utilizing a text-to-SQL dataset in a specific scenario. Finetuning in the context of LLM-based text-to-SQL can be formulated as equation~\ref{finetuning_pipeline}. Given a base model $M$ and a fine-tuning dataset $\tau=\{(q_i, d_i, gt_i)\}$, where $q_i$ represents the user query, $d_i$ is the corresponding database information, and $gt_i$ is the ground truth SQL query. The objective is to minimize the difference, as measured by the loss function $Loss$, between the predicted SQL query $M(q_i, d_i)$ and the ground truth SQL query $gt_i$.
\begin{equation}
    min_{M} = \sum_{i=0}^{|\tau|} Loss(M(q_i, d_i), gt_i)
    \label{finetuning_pipeline}
\end{equation}

\begin{figure}
    \centering
    \includegraphics[width=0.75\linewidth]{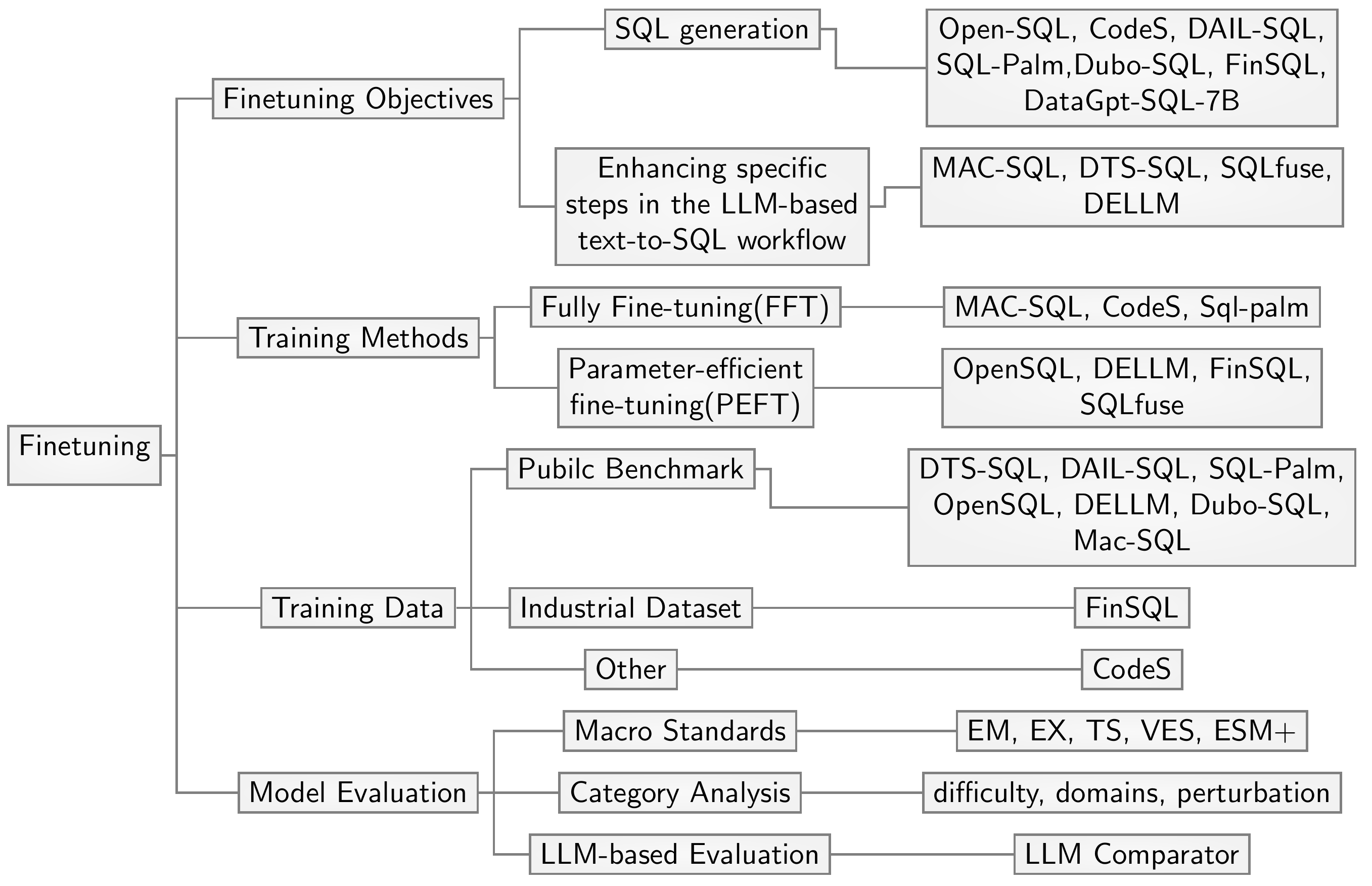}
    \caption{Taxonomy of Fine-tuning in Text-to-SQL}
    \label{fig:fine-tuning}
\end{figure}

We will structure this section into four parts, namely finetuning objectives, training methods and training data. Figure~\ref{fig:fine-tuning} presents related works classified in accordance with the aforementioned four parts.

\subsection{Finetuning Objectives}
Typically, the finetuning objective in the context of LLM-based text-to-SQL is SQL generation. That is, given the question representation and database information in specifically designed styles, models try to infer the correct SQL query. Examples of such approaches include Open-SQL\cite{opensql}, CodeS\cite{codeS}, DAIL-SQL\cite{gao2023text}, SQL-Palm\cite{sun2023sql}, Dubo-SQL\cite{Dubo}, FinSQL\cite{FinSQL} and DataGpt-SQL-7B\cite{DataGpt-SQL-7B}.

In addition to the aforementioned SQL generation objective, several studies\cite{wang2023mac, dtssql, zhang2024sqlfuseenhancingtexttosqlperformance, knowledgetosql} utilize finetuning to enhance the performance of different steps in the LLM-based text-to-SQL workflow. MAC-SQL\cite{wang2023mac} designed a three-agent architecture comprising a selector that handles schema linking, a decomposer responsible for question decomposition and SQL generation, and a refiner that deals with SQL correction. In addition to employing GPT-4\cite{achiam2023gpt} to act as the three agents, MAC-SQL\cite{wang2023mac} also explored the use of fine-tuned large language models to perform the roles of the three agents. Both DTS-SQL\cite{dtssql} and SQLfuse\cite{zhang2024sqlfuseenhancingtexttosqlperformance} explored finetuning LLMs for schema linking and sql generation. Creatively, DELLM\cite{knowledgetosql} finetuned a LLM to generate domain knowledge based on given questions and database information. This generated domain knowledge would then be integrated into prompts to assist in knowledge-intensive SQL generation.

\subsection{Training Methods}
There are two streams of training methods in LLM finetuning, namely fully fine-tuning(FFT) and parameter-efficient fine-tuning(PEFT)\cite{liu2023gpt, P-tuning, hu2021lora, dettmers2024qlora, liu2022few, lester2021power}. FFT treats all parameters in LLMs as trainable parameters while PEFT only tunes a small number of parameters, which improves training efficiency and reduce training costs. Also, according to research in \cite{jang2023exploring}, PEFT is less prone to catastrophic forgetting compared to FFT during use, which again proves its superiority.

In the context of LLM-based text-to-SQL, finetuning adopts both FFT and PEFT and among PEFT methods, LoRA\cite{hu2021lora} and QLoRA\cite{dettmers2024qlora} are most popular methods. LoRA\cite{hu2021lora}, short for Low-Rank Adaptation, freezes the pretrained model weights and injects trainable rank decomposition matrices into each layer of the Transformer architecture. In this way, LoRA significantly reduces the number of trainable parameters for downstream tasks. QLoRA\cite{dettmers2024qlora} is a finetuning method developed based on LoRA. Through a series of innovative efforts such as introducing 4-bit NormalFloat, double quantization, and Paged Optimizers, QLoRA refined the LoRA method, reducing memory usage while maintaining performance. The examples of employing FFT in the context of LLM-based text-to-SQL are Mac-SQL\cite{wang2023mac}, CodeS\cite{codeS} and Sql-palm\cite{sun2023sql} while the examples of employing LORA\cite{hu2021lora} and QLoRA\cite{dettmers2024qlora} are OpenSQL\cite{opensql}, DELLM\cite{knowledgetosql}, FinSQL\cite{FinSQL} and SQLfuse\cite{zhang2024sqlfuseenhancingtexttosqlperformance}.

\subsection{Training Data}
Due to the goal of exploring finetuning effect, most research papers\cite{wang2023mac, opensql, knowledgetosql, dtssql, gao2023text, sun2023sql, Dubo} on finetuning in the context of LLM-based text-to-SQL utilize the training set of open-source Text-to-SQL benchmarks directly. Among them, DTS-SQL\cite{dtssql}, DAIL-SQL\cite{gao2023text} and SQL-Palm\cite{sun2023sql} finetuned LLMs using Spider\cite{yu2018spider} training set. OpenSQL\cite{opensql}, DELLM\cite{knowledgetosql} and Dubo-SQL\cite{Dubo} using BIRD\cite{li2024can} training set. Mac-SQL\cite{wang2023mac} conducts experiments using both datasets. The training data may go through preprocessing (for instance, schema linking and knowledge supplementation) to be transformed into a designed representation format. 

As an example of industrial paper, FinSQL\cite{FinSQL} takes into account the unique characteristics of databases in financial applications. As public benchmarks do not consider these characteristics, FinSQL\cite{FinSQL} constructed the BULL benchmark. This benchmark is collected from the practical financial analysis business of Hundsun Technologies Inc. and includes databases for funds, stocks, and the macro economy.

In addition to the aforementioned method, CodeS\cite{codeS} sheds some light on how to construct finetuning datasets to enhance general text-to-SQL capability and how to construct finetuning datasets to adaptively transfer to databases within a new domain. For enhancing general text-to-SQL capability, CodeS\cite{codeS} collected 21.5GB of data, consisting of SQL-related data (11GB), NL-to-code data (6GB), and NL-related data (4.5GB), with the aim of enhancing both SQL generation capabilities and natural language comprehension. For adaptively transfering to databases within a new domain, CodeS\cite{codeS} collect a few genuine user queries, manually annotate corresponding SQL queries, and leverage GPT-3.5 to produce a wider set of (question, SQL) pairs in a few-shot manner, ensuring user-oriented authenticity. Meanwhile, CodeS\cite{codeS} utilize SQL templates and their question templates from text-to-SQL benchmarks. By plugging in tables, columns, and values from the databases of new domains into the templates, CodeS\cite{codeS} generate a diverse set of (question, SQL) pairs. 

\subsection{Model Evaluation}
After finetuning the model, an important step is to compare the before and after results to understand the changes in the model's performance. In the context of text-to-SQL, the common approach is to conduct a comprehensive metric analysis on the test set, such as calculating the accuracy rates of EM and EX metrics, before and after fine-tuning. However, beyond these metrics, a detailed analysis can uncover more in-depth insights. For instance, we can categorize natural language queries into several groups by considering their difficulty levels\cite{yu2018spider}, domains\cite{li2024can}, perturbation types\cite{chang2023dr.spider}, and so on. Additionally, we can classify the error types of the output results for analysis\cite{pourreza2024din,dong2023c3,zhang2024structureguidedlargelanguage,opensql}. Moreover, the emergence of LLM also opens up new possibilities for model evaluation analysis. LLM Comparator\cite{kahng2024llm} proposed a system that can visualize and evaluate the differences between two models, analyzing the differences from the "when, why, how" dimensions. Although the method of LLM Comparator\cite{kahng2024llm} is not specifically designed for the text2sql task, we can apply the query classification methods(e.g., difficulties, domains) and macro standards(e.g., EM, EX, TS\cite{zhong2020semantic}, VES\cite{li2024can}, ESM+\cite{ascoli2024esm+}) to the evaluation system of LLM Comparator\cite{kahng2024llm}, thereby migrating the evaluation system to the text2sql field.

\subsection{Current Situation}
Due to considerations of enterprise privacy information and data security, industrial text-to-SQL is more suitable for adopting the finetuning method. However, compared to prompting approaches, finetuning approaches have been relatively under-explored in the research community. Firstly, due to the superior performance of closed-source models like GPT series\cite{achiam2023gpt}, most methods centered on closed-source models can achieve better performance. Coupled with the low cost of API calls, the research on finetuning open-source models has not received as much attention as the prompt engineering method (this can be seen from the length of the prompt engineering and finetuning sections in this survey), which to some extent has led to a lag in the research of finetuning open-source model methods. Secondly, the prompt engineering method has more innovation points at the algorithmic level. In contrast, the finetuning method has fewer innovation points at the algorithmic level and more relies on progress in training techniques, base models, and the quality of training data. This also leads to a lower popularity of the finetuning for text-to-SQL method.

\begin{framed}
\textbf{Key takeaways}
\begin{itemize}
    \item Finetuning can be employed to enhance the performance of various steps in the LLM-based text-to-SQL workflow.

    \item Most research works prefer parameter-efficient fine-tuning (PEFT) over fully fine-tuning (FFT) due to PEFT's superior training efficiency and lower training costs.

    \item Open benchmarks lack the characteristics of databases in industrial applications.

    \item Leveraging large language models to generate datasets holds significant potential.
\end{itemize}
\end{framed}
\section{Model}

Whether prompt engineering or finetuning is used, the choice of base LLM is important as the implementation basis of text-to-SQL tasks is based on it. LLMs used in text-to-SQL tasks are usually based on Transformer's architecture\cite{vaswani2017attention} and adopt a decoder-only pattern suitable for open-generation tasks. We will primarily classify LLMs based on whether they are open-source. We will introduce the basic information of typical models, present the application of these models in the text-to-SQL domain, and finally analyze the development trend of basic LLMs and provide suggestions on their usage in Text2SQL tasks.

\subsection{Base Model}

After our systematic investigation, the existing LLM-based text-to-SQL works mainly use 12 closed-source LLMs and 16 open-source LLMs as base models. Some papers conduct experiments on multiple models simultaneously to compare the effects of the same approach on different models. Here we count all the models related to the approaches and experiments, rather than just the best-performing model for each method. Figure~\ref{fig:model} shows the frequency of use of these models in the paper's methods and experiments. The data is collected by identifying the specific LLMs employed for SQL generation in the experiments of each surveyed paper, including \cite{chen2023teaching,pourreza2024din,gao2023text,zhang2023act,wang2023mac,li2024pet,chess,zhang2024sqlfuseenhancingtexttosqlperformance,sui2023reboost,metasql,ren2024purplemakinglargelanguage,deasql,zhang2024structureguidedlargelanguage,r3,dong2023c3,wang2023dbcopilot,coe,liu2023divide,guo2023case,opensql,arora2023adapt,retrievalandrevision,ni2023leverlearningverifylanguagetocode,aroraageneric,cheng2022binding,chang2023prompt,fused,knowledgetosql,cogsql,ptdsql,magsql,dtssql,feathersql,basesql,mcssql,chasesql,solidsql,rslsql,esql}, all published prior to the lastest revision of this survey.

\subsubsection{Closed-source Models}

The main advantage of closed-source models lies in their large-scale pre-trained corpus and the huge parameter size. Taking GPT-4\cite{achiam2023gpt} as an example, it is pre-trained on about 13T tokens (including text data and code data) and has about 1.8 trillion (1800B) parameters across 120 layers. GPT-4 is also the most common base model in LLM-based text-to-SQL work\cite{chen2023teaching,pourreza2024din,gao2023text,zhang2023act,wang2023mac,li2024pet,chess,zhang2024sqlfuseenhancingtexttosqlperformance,sui2023reboost,metasql,ren2024purplemakinglargelanguage,deasql,r3,zhang2024structureguidedlargelanguage}. Although slightly inferior to GPT-4, GPT-3.5 series also act as a strong baseline model for most text-to-SQL methods \cite{chen2023teaching,dong2023c3,gao2023text,zhang2023act,wang2023dbcopilot,chess,coe,liu2023divide,tai2023exploring,sui2023reboost,metasql,ren2024purplemakinglargelanguage,deasql,zhang2024structureguidedlargelanguage,r3,opensql,arora2023adapt}. As a descendant of GPT-3, the closed-source LLM CodeX series\cite{chen2021evaluating} which focuses on code generation is pre-trained on billions of lines of code and has more than 14K Python code memory. CodeX has been used in many text-to-SQL approaches\cite{ni2023leverlearningverifylanguagetocode,chen2023teaching,pourreza2024din,aroraageneric,cheng2022binding,zhang2024structureguidedlargelanguage,guo2023prompting,tai2023exploring} due to its expertise in many programming languages. In addition, text-davinci-003 model and PaLM-2\cite{anil2023palm} are also used by some methods\cite{liu2023divide,retrievalandrevision,arora2023adapt,zhang2024structureguidedlargelanguage,gao2023text} in the text-to-SQL domain. Since closed-source large models have large parameters and are difficult to deploy independently, most methods or experiments in the literature work use API calls to access these models. However, as of the time of writing this article, there is no closed-source LLM specifically for SQL code generation. There are two possible reasons: First, SQL involves privacy information in different business data processing, so it is difficult to directly obtain a certain scale of pre-training data; second, the generalization of the language model specifically generated for SQL may not be as good as the general code generation model. This part can be a direction for future exploration.

\subsubsection{Open-source Models}

Unlike closed-source models, open-source models are often deployed or fine-tuned in the private domain by users with certain hardware support due to their appropriate parameter scale. Usually, open-source LLMs have versions of different model sizes. Generally speaking, the more model parameters, the stronger the model generation capability\cite{gao2023text}. The Llama 3 series\cite{llama3} commonly used in text-to-SQL is divided into two types: 8B parameter size and 70B parameter size. In addition, the Code Llama series\cite{Code-llama} has several parameter sizes of 7B, 13B, 34B, and 70B, and its performance in code generation and completion in languages such as Python and C++ exceeds that of the non-dedicated model Llama 2\cite{gao2023text}. Deepseek\cite{deepseekr1} and QWen\cite{qwen25} also use version numbers to indicate their different model types with various parameter sizes. Among the open-source large models, SQLCoder\cite{sqlcoder} is a model family dedicated to text-to-SQL tasks. Its 70B parameter version and 15B parameter version respectively surpass the performance of GPT-4 and GPT-3.5-turbo on a new dataset based on SPIDER\cite{yu2018spider} with novel schemas not seen in training. Most text-to-SQL methods based on prompt engineering that use open-source models often view them as baselines for experiments. For example, DAIL-SQL\cite{gao2023text} compared Llama, Llama 2, Coda Llama, Alpaca\cite{alpaca}, and Vicuna models\cite{vicuna} in the experiment. The latter two are general models fine-tuned based on Llama 7B and Llama 13B respectively. PET-SQL\cite{li2024pet} compared Code Llama, SQLCoder, InternLM\cite{cai2024internlm2}, and SenseChat\cite{SenseChat} in the experiment, while DEA-SQL\cite{deasql} selected Llama 2, Code Llama, and WizardCoder\cite{xu2023wizardlm} as the base models of its method. In the work using the finetune method, LEVER\cite{ni2023leverlearningverifylanguagetocode} fine-tuned InCoder\cite{fried2022incoder} and CodeGen\cite{nijkamp2022codegen}. Open-SQL\cite{opensql} fine-tuned Llama 2 and Code Llama in the text-to-SQL domain. The effect of Code Llama after fine-tuning on the BIRD dataset has exceeded that of the closed-source LLM GPT-4. CHESS\cite{chess} uses the Llama 3 70B model and the fine-tuned DeepSeek Coder\cite{guo2024deepseek} model, and its effect on the BIRD-dev dataset is comparable to that of the methods based on closed-source LLMs.

\begin{figure}
    \centering
    \includegraphics[width=\linewidth]{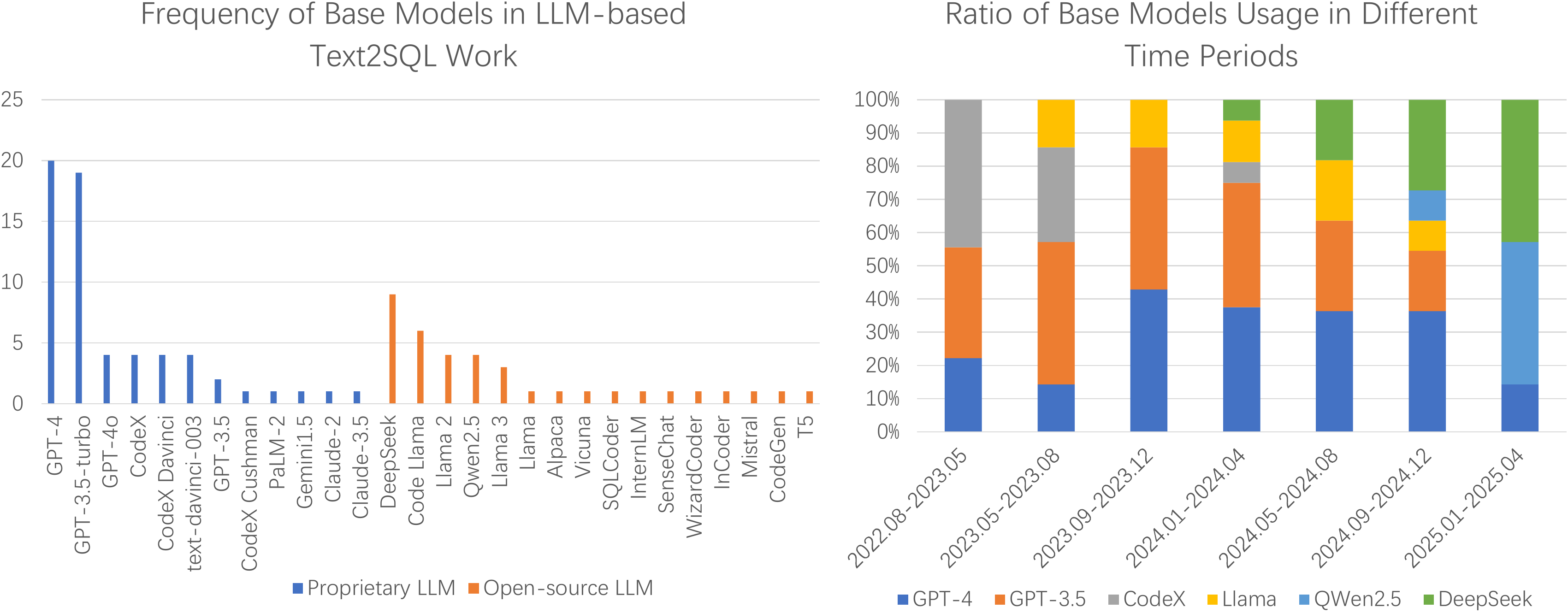}
    \caption{Frequency of LLM usage in Text2SQL research works from August 2022 to April 2025 (left). Trend of base model series usage in text-to-SQL tasks (right). }
    \label{fig:model}
\end{figure}

\subsection{Trend}

Figure~\ref{fig:model} shows the trend of the usage frequency of the base models in the LLM-based text-to-SQL methods and experiments over time, as well as the time when each base model series was launched. It can be found that within the scope of closed-source LLMs, the emergence of GPT-4\cite{achiam2023gpt} (in March 2023) has made most of the work prefer to use the GPT series models rather than CodeX series; within the scope of open-source LLMs, most of the work will choose Deepseek\cite{deepseekr1}, Llama, Code Llama , and QWen\cite{qwen25} series of base models. As time went by, the frequency of use of closed-source models increased in 2023 but decreased in 2024, while the open-source models were the opposite. One possible reason is that the growing parameter size in open-source models makes them improve task performance while maintaining privacy and fine-tuning abilities. In the future, as the cost of LLMs decreases, the most powerful closed-sourced LLM and the most powerful open-sourced LLM will become dominant in text-to-SQL tasks.

\begin{framed}
\textbf{Key takeaways}
\begin{itemize}
    \item In current text-to-SQL tasks, the advantage of closed-source base models is that they are ready to use and have powerful code-generation capabilities. A combination of closed-source models and prompt engineering methods is suitable for users with insufficient hardware support.
    \item The advantages of using open-source base models are independent deployment, privacy, and the ability to fine-tune domains. Users with hardware support can combine open-source models with fine-tuning for in-domain text-to-SQL tasks.
    \item Existing LLM-based text-to-SQL work favors the GPT series among closed-source models and the DeepSeek, Llama, Code Llama, and Qwen series among open-source models.
    \item Until now, open-source models have been used more and more frequently and their usage is equally matched to that of closed-source models in the research aspect.
\end{itemize}
\end{framed}
\section{Analysis}
To further investigate the practical value and applicability of each method, we conduct a comparative analysis of experimental results from existing studies.

\subsection{Benchmark Selection}
We evaluate performance on three widely used benchmarks in Text-to-SQL research: Spider 1.0\cite{yu2018spider}, BIRD\cite{li2024can}, and Spider 2.0\cite{spider2}, each representing key challenges in the field.

Spider 1.0\cite{yu2018spider} remains the most established benchmark, serving as the primary dataset for evaluating Text-to-SQL model performance. Its standardized schema and well-defined complexity levels make it a cornerstone for comparative studies. BIRD\cite{li2024can} introduces real-world challenges, such as handling noisy and incomplete database values, as well as incorporating external knowledge to bridge natural language questions with database content. This dataset significantly narrows the gap between academic research and practical applications. Spider 2.0\cite{spider2} extends the scope to enterprise-level workflows, featuring complex multi-database environments (both cloud and local), diverse SQL dialects, and a broad range of operations—from data transformation to analytical queries.

Overall, these datasets demonstrate a clear trend toward greater realism, with SQL queries becoming more intricate and challenging, closely mirroring industrial use cases.

\subsection{Evaluation Methodology}

To ensure a fair comparison, we extract results from the official leaderboards of Spider1.0\cite{yu2018spider}, BIRD\cite{li2024can}, and Spider 2.0\cite{spider2}, adhering to the following selection criteria: 1. Only approaches leveraging LLMs are considered; 2. A maximum of two best-performing LLMs per method are included; 3. Methods based on models released prior to 2022 (pre-ChatGPT) are excluded to focus on recent advancements; 4. Methods relying on undocumented models (e.g., closed systems lacking technical details or API access) are disregarded; 5. Non-reproducible approaches (e.g., those without open-source code, published papers, or sufficient implementation details) are excluded to maintain fairness and practical utility.

\begin{figure}
    \centering
    \includegraphics[width=0.99\linewidth]{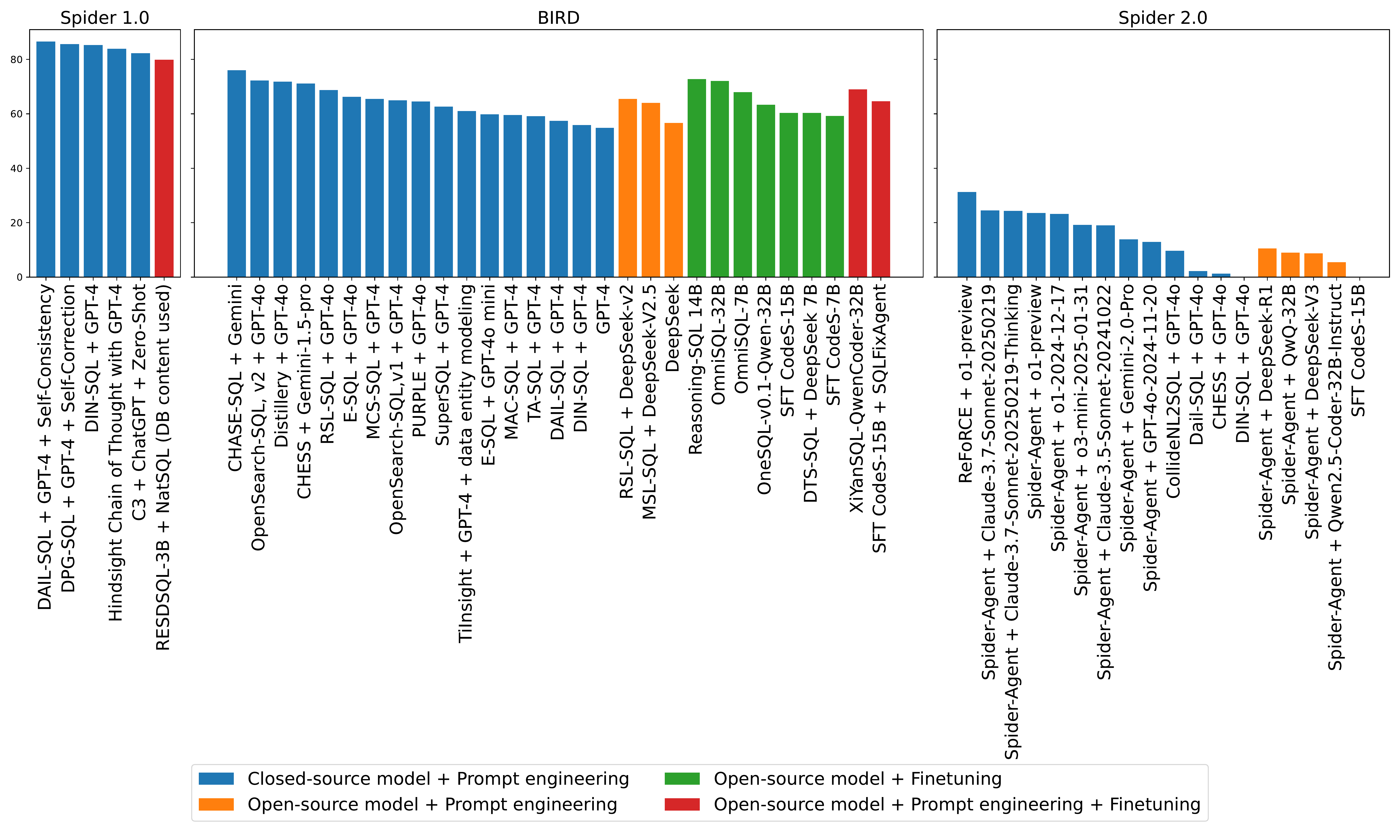}
    \caption{Execution accuracy of different combinations of methods and LLMs on the test sets of the Spider 1.0\cite{yu2018spider}, BIRD\cite{li2024can}, and Spider 2.0\cite{spider2} benchmarks.}
    \label{fig:leardboard}
\end{figure}

\subsection{Results}

We adopt the EX metric and the results are shown in Fig.~\ref{fig:leardboard}.

\textbf{Spider 1.0}. All methods included in our analysis represent typical LLM-based Text-to-SQL approaches, with most achieving pass rates exceeding 80\%. This demonstrates that modern large language models have effectively solved basic Text-to-SQL tasks on the Spider benchmark, confirming their competence in handling fundamental semantic parsing challenges.

\textbf{BIRD}. At the model level, both powerful closed-source and open-source LLMs combined with prompting strategies, as well as smaller open-source LLMs enhanced through fine-tuning, demonstrate competitive, state-of-the-art performance. This suggests that the performance gap between closed-source and open-source models has narrowed significantly and compact models enhanced via fine-tuning can show strong potential to match the capabilities of much larger LLMs.

At the methodological level, our comparison focuses on approaches using the same underlying LLMs to ensure fairness. Although limited experimental data makes it difficult to pinpoint the exact reasons for performance differences between methods, several distinctive characteristics can be observed in the most effective approaches. In particular, we highlight OpenSearch-SQL\cite{OpenSearch-SQL} (the top-performing method under GPT-4o and second-best under GPT-4), Distillery\cite{Distillery} (second-best under GPT-4o), and MCS-SQL\cite{mcssql} (top-performing under GPT-4).

\textit{Characteristic 1: Inference-time Scaling.} These top-performing methods leverage advanced inference-time strategies. OpenSearch-SQL\cite{OpenSearch-SQL} incorporates multi-SQL generation, correction mechanisms, and self-consistency checks. Distillery\cite{Distillery} employs iterative correction combined with self-consistency, while MCS-SQL\cite{mcssql} adopts multiple prompts for schema linking and SQL generation, thereby exploring a broader search space for potential answers. In contrast, baseline methods such as vanilla DIN-SQL\cite{pourreza2024din} rely on single-step SQL correction, and vanilla DAIL-SQL\cite{gao2023text} lacks any correction or self-consistency components.

\textit{Characteristic 2: Enhanced Schema Linking.} Distillery\cite{Distillery} investigates the impact of schema linking errors and observes that modern LLMs are increasingly tolerant of irrelevant columns. Consequently, it recommends including as many columns as possible when context length permits. MCS-SQL\cite{mcssql}, on the other hand, enhances schema linking through its inference-time scaling strategy, which broadens the exploration space and improves the likelihood of generating correct SQL queries.

\textbf{Spider 2.0}. The Spider 2.0\cite{spider2} evaluation provides valuable insights into the capabilities and limitations of LLM-based Text-to-SQL systems. The performance of Spider-Agent\cite{spider2} across different large language models (LLMs) highlights that, although open-source models have made significant progress, they still lag behind proprietary alternatives. The evaluation also shows that reasoning-enhanced models consistently deliver stronger performance. Furthermore, we observe that non-agent approaches significantly underperform compared to agent-based methods. This may be attributed to the complexity of real-world Text-to-SQL tasks, which often involve diverse SQL dialects, complex syntax and functions, nested columns, the integration of external knowledge, and the need to interpret ambiguous user requests. These challenges underscore the importance of agentic approaches, which allow LLMs to dynamically interact with databases, receive feedback, and iteratively plan their actions. Notably, REFORCE\cite{ReFoRCE}, a recent agent-based method, outperforms standard Spider agents by incorporating advanced techniques such as table compression for context optimization, structured output formatting, iterative column exploration for better schema understanding, and a novel self-refinement pipeline that combines parallelized voting with CTE-based resolution. Despite these promising developments, it is important to note that the current state-of-the-art method achieves only a 31.26\% execution accuracy, underscoring a significant gap between academic advancements and real-world deployment.

\begin{framed}
\textbf{Key takeaways}
\begin{itemize}
    \item LLMs excel on Spider 1.0, showing that foundational Text-to-SQL tasks are largely solved by modern models.
    
    \item Effective inference-time strategies and schema linking may significantly boost performance.
    
    \item Spider 2.0 reveals real-world gaps, where agent-based methods outperform others, but overall execution accuracy remains low, highlighting the need for more robust, deployable solutions.
\end{itemize}
\end{framed}
\section{Future Direction}

Although LLM technology has significantly improved text-to-SQL tasks, there are still many difficulties in developing high-quality text-to-SQL parsers. In this section, we will first conduct a review of the error analysis of research approaches. Subsequently, we will introduce the challenges faced by existing solutions and some promising future development directions for LLM-based text-to-SQL tasks.

\subsection{Error Analysis Overview}

Some existing methods \cite{pourreza2024din,dong2023c3,zhang2024structureguidedlargelanguage,opensql} have conducted error analysis on the test results of SPIDER and BIRD datasets. Their error classification method is based on the components that produce errors in the generated SQL. There are five major categories: schema-linking, join, nested, group by, and others (Open-SQL\cite{opensql} categorizes "group by" into others). Each major category also has subcategories such as wrong column, wrong table, and wrong condition. Among them, the average error rate in schema-linking is the highest, at 29\%-49\%, the error rate near \textit{JOIN} is 21\%-26\%, and the error rate of \textit{GROUP BY} and nested is less than 20\%. This shows that schema-linking is still the module that needs to be improved most in the current text-to-SQL task. The join connection and syntax problems of data tables also have considerable challenges. In \cite{liu2024survey}, the errors listed by DIN-SQL are classified more finely for SQL keywords, involving keywords such as \textit{SELECT}, \textit{FROM}, \textit{WHERE}, \textit{ORDER BY}, \textit{GROUP BY}, and more fine-grained types such as error, redundancy, and insufficiency. In addition, when performing error analysis, MAC-SQL\cite{wang2023mac} found that the gold answer in the SPIDER\cite{yu2018spider} and BIRD\cite{li2024can} benchmarks also had 20\%-30\% errors, demonstrating the necessity of data cleaning.

\subsection{Practical Challenges and Directions}

\subsubsection{Privacy Concern}

ChatGPT and GPT-4\cite{achiam2023gpt} demonstrate impressive capabilities, acknowledged as the most powerful LLMs. However, concerns over privacy issues often arise while using API call models in production scenarios, because of the potential leakage of private information through feeding prompts to the API. Private deployment of open-sourced LLM and fine-tuning it can serve as a solution to this problem while maintaining performance. However, the dirty data, possible compromise in LLM's other capabilities, and catastrophic forgetfulness are blocking the performance gain of finetuning in production scenarios. Extracting high-quality training data may help handle the above private fine-tuning issue.

\subsubsection{Complex Schema and Insufficient Benchmark}

Table schemas are much more complex than current research benchmarks in many real-world text-to-SQL tasks. One of Microsoft's internal financial data warehouse\cite{NL2SQLisnot} encompasses 632 tables with over 4000 columns and 200 views with more than 7400 columns. The text-to-SQL task on this database shows some prominent issues: difficulty in schema linking, diversion of LLM's attention due to a torrent of tokens, and long inference time. To address schema-linking challenges, recent advancements provide promising directions. Some approaches\cite{ReFoRCE, spider2} leverage agent-based methods, such as ReAct~\cite{yao2022react}, to interactively explore database structures, while others\cite{mcssql} employ self-consistency techniques to reduce schema-linking errors. Beyond these methods, emerging developments in dynamic graph attention mechanisms\cite{DBLP:journals/eswa/ChenZD25} present promising opportunities. Specifically, adaptive graph-based representations can be constructed to model the relationships among tables, columns, and queries.

Current standard datasets can not model the complexity of the above Microsoft's warehouse. Spider\cite{yu2018spider} only contains simple table schemas. BIRD\cite{li2024can} emulates real-world scenarios, but the scale is far not enough. Spider 2.0 \cite{spider2} provides a good example to incorporate common local or cloud database systems such as BigQuery and Snowflake, showing the potential direction of real-world usage of Text-to-SQL benchmarks.

\subsubsection{Domain Knowledge}

The training of large language models (LLMs) on extensive corpora equips them with broad general knowledge, forming the basis for their strong text-to-SQL capabilities. However, real-world applications such as data analysis and batch processing demand domain-specific knowledge, including industry terminology, abbreviations, and jargon. Without such knowledge, LLMs often misinterpret or inaccurately respond to queries. Domain adaptation can be achieved through prompt engineering and fine-tuning, though both face significant challenges. Prompt engineering, often relying on Retrieval-Augmented Generation (RAG)\cite{gao2023retrieval}, requires a high-quality structured knowledge base. In practice, domain knowledge is typically embedded in noisy, unstructured documents, complicating knowledge base construction. Furthermore, similarity-based retrieval may introduce irrelevant information, degrading model performance. Fine-tuning, while embedding domain knowledge directly into model parameters, risks catastrophic forgetting and limits future adaptability; modifying or updating embedded knowledge often necessitates repeated, costly retraining.

\subsubsection{Autonomous Agents}

Building on the ReAct framework \cite{yao2022react}, it is now feasible to develop autonomous agents powered by LLMs. In such architectures, the LLM serves as the agent’s core, orchestrating interactions with environments, humans, and other agents, while leveraging short- and long-term memory to iteratively refine task execution. The motivation for applying agents to text-to-SQL is from how humans naturally compose SQL: human SQL construction involves trial-and-error cycles of execution and revision, a workflow well-suited for agent-based modeling. Notably, REFORCE \cite{ReFoRCE} and Spider 2.0 \cite{spider2} are the first frameworks to explicitly adopt autonomous agents for text-to-SQL tasks. Results from the Spider 2.0 leaderboard show that traditional non-agent methods \cite{pourreza2024din, gao2023text, chess} underperform compared to these agent-based approaches, highlighting the potential of LLM-driven agents in this domain.

\subsubsection{Data Governance}

Two key limitations in current text-to-SQL datasets are ambiguity and semantic mismatch\cite{NL2SQLisnot}. Ambiguity arises when multiple semantically distinct SQL queries yield valid answers to the same natural language question, complicating the evaluation of model correctness. Semantic mismatch occurs when user queries reflect intents that cannot be answered by the underlying database, either partially or entirely. In the context of LLM-based text-to-SQL systems, data governance offers three major benefits: (1) structuring domain knowledge enhances RAG effectiveness and improves alignment between natural language queries and executable logic, thereby reducing ambiguity; (2) improving training data quality strengthens LLMs' domain-specific capabilities during fine-tuning; and (3) refining benchmark datasets to address semantic mismatch enables more robust evaluation of practical system performance.

\section{Conclusion}

This paper offers a comprehensive review of the utilization of Large Language Models (LLMs) in Text-to-SQL tasks. We first enumerate classic benchmarks and evaluation metrics. For the two mainstream methods, prompt engineering and finetuning, we introduce a comprehensive taxonomy and offer practical insights into each subcategory. We present an overall analysis of the above methods and various models evaluated on well-known datasets and extract some insights. Finally, we discuss the challenges and future directions in this field.



\end{document}